\documentclass[preprint,authoryear,1p]{elsarticle}

\usepackage[ruled,linesnumbered]{algorithm2e}
\usepackage{graphicx}
\usepackage{times,amsmath,newlfont,amssymb}
\usepackage[dvips,counterclockwise,figuresright]{rotating}
\usepackage{natbib}
\usepackage{epsfig}
\usepackage{multirow}

\DeclareMathOperator*{\argmin} {argmin}
\DeclareMathOperator*{\mymax} {max}
\DeclareMathOperator*{\mymin} {min}

\newcommand{\calP}{\mathcal{P}}

\newcommand{\calS}{\mathcal{S}}

\newcommand{\lpg}{\mbox{\sc lpg}}
\newcommand{\ddkset}{$d${\sc distance}$k${\sc set}}

\def\mvp{\vspace*{-0.1in}}

\begin{document}

\journal{Artificial Intelligence}

\begin{frontmatter}
\title{Planning with Partial Preference Models}

\author[asu]{Tuan Nguyen\corref{cor1}\corref{cor2}}
\ead{natuan@asu.edu}

\author[parc]{Minh Do}
\ead{minh.do@parc.com}

\author[brescia]{Alfonso Gerevini}
\ead{gerevini@ing.unibs.it}

\author[freeuni]{Ivan Serina}
\ead{ivan.serina@unibz.it}

\author[ibm]{Biplav Srivastava}
\ead{sbiplav@in.ibm.com}

\author[asu]{Subbarao~Kambhampati}
\ead{rao@asu.edu}

\cortext[cor1]{Corresponding author.}

\cortext[cor2]{Authors listed in alphabetical order, with the exception of
  the first and the last.}

\address[asu]{
Department of Computer Science and Engineering, Arizona State University
Brickyard Suite 501, 699 South Mill Avenue, Tempe, AZ 85281, USA}

\address[parc]{
Embedded Reasoning Area, Palo Alto Research Center 3333 Coyote Hill Road, Palo Alto, CA 94304, USA}

\address[brescia]{Dipartimento di Elettronica per l'Automazione, Università degli Studi di Brescia, Via Branze 38, I-25123 Brescia, Italy}

\address[freeuni]{Free University of Bozen-Bolzano, Viale Ratisbona, 16, I-39042 Bressanone, Italy}

\address[ibm]{IBM India Research Laboratory, New Delhi and Bangalore, India}



\begin{abstract}

  Current work in planning with preferences assume that the user's
  preference models are completely specified and aim to search for a
  single solution plan. In many real-world planning scenarios, however,
  the user probably cannot provide any information about her desired
  plans, or in some cases can only express partial preferences. In such
  situations, the planner has to present not only one but a set of plans
  to the user, with the hope that some of them are similar to the
  plan she prefers. We first propose the usage of different measures to
  capture quality of plan sets that are suitable for such scenarios:
  domain-independent distance measures defined based on plan elements
  (actions, states, causal links) if no knowledge of the user's preferences is
  given, and the \emph{Integrated Convex Preference} measure in case the
  user's partial preference is provided. We then investigate various
  heuristic approaches to find set of plans according to these measures,
  and present empirical results demonstrating the promise of our
  approach.\footnote{This work is an
    extension of the work presented in
    \cite{srivastava07} and \cite{nguyen2009planning}.}

\end{abstract}


\end{frontmatter}

\section{Introduction}
\label{sec:intro}
\nocite{}
\noindent

Most work in automated planning takes as input a complete specification
of domain models and/or user preferences and the planner searches for a
single solution satisfying the goals, probably optimizing some objective
function. In many real world planning scenarios, however, the user's
preferences on desired plans are either unknown or at best partially
specified (c.f. \cite{model-lite}). In such cases, the planner's job
changes from finding a single optimal plan to finding a set of
representative solutions (``options'') and presenting them to the user
with the hope that she can find one of them desirable. As an example, in
adaptive web services composition, the causal dependencies among some
web services might change at the execution time, and as a result the web
service engine wants to have a set of diverse plans/compositions such
that if there is a failure while executing one composition, an
alternative may be used which is less likely to be failing
simultaneously \citep{chafle2006adaptation}. However, if a user is
helping in selecting the compositions, the planner could be first asked
for a set of plans that may take into account the user's trust
in some particular sources and when she selects one of them, it is next
asked to find plans that are similar to the selected one. The
requirement of searching for a set of plans is also considered in
intrusion detection \citep{boddy2005course} where a security analysis
needs to analyze a set of attack plans that might be attempted by a
potential adversary, given limited (or unknown) information about the
adversary's model (e.g., his goals, capabilities, habits, ...), and
the resulting analyzed information can then be used to set up defensive
strategies against potential attacks in the future. Another example can
be found in \cite{gui-test-plans} in which test cases for graphical user
interfaces (GUIs) are generated as a set of distinct plans, each
corresponding to a sequence of actions that a user could perform, given
the user's unknown preferences on how to interact with the GUI to
achieve her goals. The capability of synthesizing multiple plans would
also have potential application in case-based planning
(e.g., \cite{serina2010kernel}) where it is important to have a plan set
satisfying a case instance. These plans can be different in terms of
criteria such as resources, makespan and cost that can only be specified in the
retrieval phase. In the problem of travel planning for individuals of a
city in a distributed manner while also optimizing public resource (e.g., road,
traffic police personel), the availability of a number of plans for each
person's goals could make the plan merging phase easier and reduce the
conflicts among individual plans.

In this work, we investigate the problem of generating \emph{a set of
  plans} in order to deal with planning situations where the preference
model is not completely specified. In particular, we consider the following scenarios:

\begin{itemize}

\item Even though the planner is aware that the user has some
  preferences on solution plans, it is not provided with any of that
  knowledge.

\item The planner is provided with incomplete knowledge of the user's
  preferences. In particular, the user is interested in some plan
  \emph{attributes} (such as the duration and cost of a flight, or
  whether all packages with priority are delivered on time in a logistic
  domain), each with different but unknown degree of importance
  (represented by \emph{weight} or \emph{trade-off} values). Normally,
  it is quite hard for a user to indicate the exact trade-off values, but
  instead more likely to determine that one attribute is more (or less)
  important than some others---for instance, a bussinessman would
  consider the duration of a flight much more important
  than its cost. Such kind of incomplete preference specification could
  be modeled with a probability distribution of weights
  values\footnote{Even if we do not have any special knowledge about
    this probability distribution, we can always start by initializing
    it to be uniform, and gradually improve it based on interaction with
    the user.}, and is therefore assumed to be given as an input
  (together with the attributes) to the planner.

\end{itemize}

\noindent
Even though, in principle, the user would have a better chance to find
her desired plan from a larger plan set, there are two problems to
consider---one computational, and other comprehensional. The
computational problem is that synthesis of a single plan is often quite
costly already, and therefore it is even more challenging to search for
a large plan set. Coming to the second problem, it is unclear that the user
will be able to inspect a large set of plans to identify the plan she
prefers. What is clearly needed, therefore, is the ability to generate a
set of plans, among all sets of \emph{bounded} (small) number of plans,
with the highest chance of including the user's preferred plan. An
immediate challenge is formalizing what it means for a \emph{meaningful}
set of plans, in other words what the \emph{quality measure} of plan
sets should be given an incomplete preference specification.

We propose different quality measures for the two scenarios listed above. In
the extreme case when the user could not provide any knowledge
of her preferences, we define a spectrum of distance measures between two plans based on their
syntactic features in
order to define the \emph{diversity} measure of plan sets. These
measures can be used regardless of the user's preference, and
by maximizing the diversity of a plan set we increase the chance that the
set is uniformly distributed in the unknown preference space, and
therefore likely contains a plan that is close to a user's desired one.

This measure can be further refined when some knowledge of the user's
preferences is provided. As mentioned above, we assume that the user's
preference is specified by a convex combination of plan attributes, and
incomplete in the sense that the distribution of trade-off weights is
given, not their exact values. The whole set of best plans (i.e. the
ones with the best value function) can be pictured as the lower
convex-hull of the Pareto set on the attribute space. To measure the
quality of any (bounded) set of plans on the whole optimal set, we adapt
the idea of \emph{Integrated Preference Function} (IPF) \citep{ipf}, in
particular its special case \emph{Integrated Convex Preference} (ICP).
This measure was developed in the Operations Research (OR) community in
the context of multi-criteria scheduling, and is able to associate a
robust measure of representativeness for any set of solution schedules
\citep{fowler2005evaluating}.

Armed with these quality measures, we can then formulate the problem of
planning with partial preference models as finding a bounded set
of plans that has the best quality value. Our next contribution
therefore is to investigate effective approaches for using quality
measures to bias a planner's search to find a high quality plan set
efficiently. For the first scenario when the preference specification is
not provided, two representative state-of-the-art planning approaches
are considered.  The first, {\sc gp-csp} \citep{gp-csp}, typifies the
issues involved in generating diverse plans in bounded horizon
compilation approaches, while the second, {\sc lpg} \citep{lpg:JAIR03},
typifies the issues involved in modifying the heuristic search planners.
Our investigations with {\sc gp-csp} allow us to compare the relative
difficulties of enforcing diversity with each of the three different distance
measures (elaborated in later section). With {\sc lpg}, we find that the proposed quality measure
makes it more effective in generating plan set over large problem
instances. For the second case when part of the user's preferences is
provided, we also present a spectrum of approaches for solving this
problem efficiently. We implement these approaches on top of Metric-LPG
\citep{metric-lpg}. Our empirical evaluation compares these approaches
both among themselves as well as against the methods for generating
diverse plans ignoring the partial preference information, and the
results demonstrate the promise of our proposed solutions.

\begin{figure*}[t]
\centering
\epsfig{file=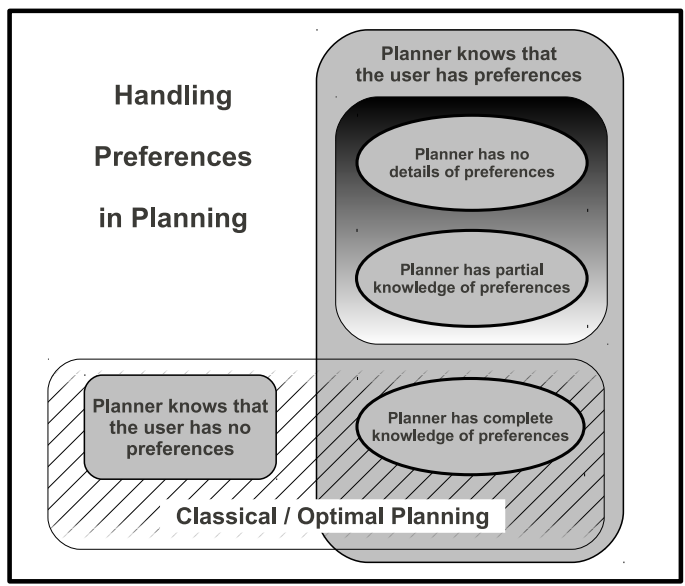,width=3.5in}
\caption{An overview picture of planning with respect to knowledge of user's preferences.}
\label{fig:overview}
\end{figure*}

Our work can be considered as a complement to current research in
planning with preferences, as shown in Figure~\ref{fig:overview}. Under
the perspective of planning with preferences, most current work in
planning synthesize a single solution plan, or a single best one, in
situations where user has no preferences, or a complete knowledge of
preferences is given to the planner. On the other hand, we address the
problem of synthesizing a set of plans when knowledge of user's
preferences is either completely unknown or partially specified.

The paper is organized as follows.
Section~\ref{sec:background-notations} gives fundamental concepts in
preferences, and formal notations. In Section~\ref{sec:quality-measures},
we formalize quality measures of plan set in the two scenarios.
Sections~\ref{sec:planning-no-preference} and
\ref{sec:planning-partial-preference} discuss our various heuristic
approaches to generate plan sets, together with the experimental
results. We discuss related work in Section~\ref{sec:related-work},
future work and conclusion in Section~\ref{sec:future}.

\section{Background and Notation}
\label{sec:background-notations}
\noindent
Given a planning problem with the set of solution plans $\calS$, a user
preference \emph{model} is a transitive, reflextive relation in $\calS
\times \calS$, which defines an ordering between two plans $p$ and $p'$
in $\calS$. Intuitively, $p \preceq p'$ means that the user prefers $p$
at least as much as $p'$. Note that this ordering can be either partial
(i.e. it is possible that neither $p \preceq p'$ nor $p' \preceq p$
holds---in other words, they are incomparable), or total (i.e. either $p
\preceq p'$ or $p' \preceq p$ holds). A plan $p$ is considered more
preferred than a plan $p'$, denoted by $p \prec p'$, if $p \preceq p'$,
$p' \not\preceq p$, and they are equally preferred if $p \preceq p'$ and
$p' \preceq p$. A plan $p$ is an optimal (i.e., most preferred) plan if
$p \preceq p'$ for any other plan $p'$. A plan set $\calP \subseteq
\calS$ is considered more preferred than $\calP' \subseteq \calS$,
denoted by $\calP \prec \calP'$, if $p \prec p'$ for any $p \in \calP$
and $p' \in \calP'$, and they are incomparable if there exists $p \in
\calP$ and $p' \in \calP'$ such that $p$ and $p'$ are incomparable.

\begin{figure*}[t]
\centering
\epsfig{file=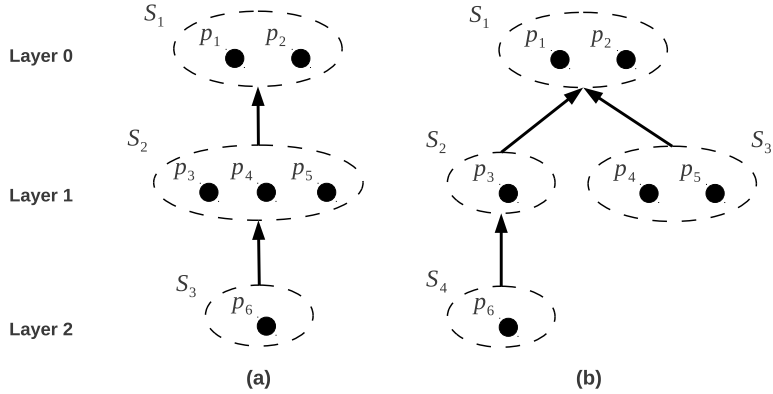,width=4in}
\caption{The Hasse diagrams and layers of plan sets implied by two preference
  models. In (a), $\calS_1 \prec \calS_2 \prec \calS_3$, and any two
  plans are comparable. In (b), on the other hand, $\calS_1 \prec
  \calS_2 \prec \calS_4$, $\calS_1 \prec \calS_3$, and each plan in
  $\calS_3$ is incomparable with plans in $\calS_2$ and $\calS_4$.}
\label{fig:diagram}
\end{figure*}

The ordering $\preceq$ implies a partition of $\calS$ into disjoint plan
sets (or \emph{classes}) $\calS_0$, $\calS_1$, ...  ($\calS_0 \cup
\calS_1 \cup ... = \calS$, $\calS_i \cap \calS_j = \emptyset$) such that
plans in the same set are equally preferred, and for any set $\calS_i$,
$\calS_j$, either $\calS_i \prec \calS_j$ , $\calS_j \prec \calS_i$, or
they are incomparable. The partial ordering between these sets can be
represented as a Hasse diagram~\citep{birkhoff1948lattice} where the
sets are vetices, and there is an (upward) edge from $\calS_j$ to
$\calS_i$ if $\calS_i \prec \calS_j$ and there is not any $\calS_k$ in
the partition such that $\calS_i \prec \calS_k \prec \calS_j$. We denote
$l(\calS_i)$ as the ``layer'' of the set $\calS_i$ in the diagram,
assuming that the most preferred sets are placed at the layer 0, and
$l(\calS_j) = l(\calS_i) + 1$ if there is an edge from $\calS_j$ to
$\calS_i$. A plan in a set at a layer with the smaller value, in
general, is either more preferred than or incomparable with ones at
high-value layers.\footnote{If $\preceq$ is a total ordering, then plans
  at smaller layer is more preferred than ones at higher layer.}
Figure~\ref{fig:diagram} show examples of Hasse diagrams representing a
total and partial preference ordering between plans.

When the preference model is explicitly specified, answering queries
such as comparing two plans, finding a most preferred (optimal) plan
becomes an easy task. This is possible, however, only if the set of
plans is small and known upfront. Many preference \emph{languages},
therefore, have been proposed to represent the relation $\preceq$ in a
more compact way, and serve as starting points for \emph{algorithms} to
answer queries. Most preference languages fall into the following two
categories:

\begin{itemize}

\item Quantitative languages define a \emph{value function} $V: \calS
  \rightarrow R$ which assigns a real number to each plan, with a
  precise interpretation that $p \preceq p' \iff V(p) \leq V(p')$.
  Although this function is defined differently in many languages, at a
  high level it combines the user's preferences on various aspects of
  plan that can be measured quantitatively. For instance, in the context
  of decision-theoretic planning \citep{boutilier1999:mdp}, the value
  function of a policy is defined as the expected rewards of states that
  are visited when the policy executes. In partial satisfaction
  (over-subcription) planning (PSP)
  \citep{smith2004choosing,van2004psp}, the quality of plans is defined
  as its total rewards of soft goals achieved minus its total action
  costs. In PDDL2.1 \citep{pddl2.1}, the value function is an arithmetic
  function of numerical fluents such as plan makespans, fuel used etc.,
  and in PDDL3 \citep{gerevini2009deterministic} it is enhanced with
  individual preference specification defined as formulae over state
  trajectory using linear temporal logic (LTL) \citep{pnueli1977:ltl}.


\item Qualitative languages provide qualitative statements that are more
  intuitive for lay users to specify. A commonly used language of this
  type is \emph{CP-networks} \citep{boutilier2004cp}, where the user can
  specify her preference statements on values of plan attributes,
  possibly given specification of others (for instance, ``Among tickets
  with the same prices, I prefer airline A to airline B.''). Another
  example is \emph{LPP} \citep{bienvenu2006planning} in which the
  statements can be specified using LTL formulae, and possibly being
  aggregated in different ways.

\end{itemize}

\begin{figure*}[t]
\centering
\epsfig{file=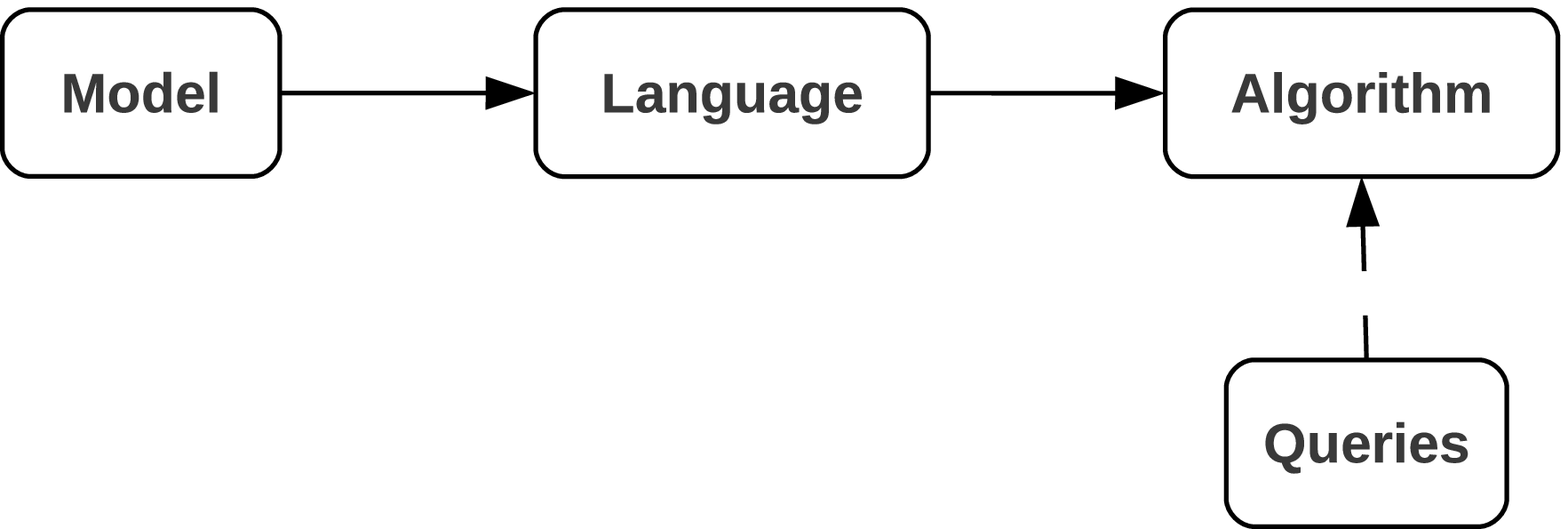,width=3in}
\caption{The metamodel \citep{brafman2009preference}.}
\label{fig:metamodel}
\end{figure*}

Figure~\ref{fig:metamodel} shows the conceptual relation of preference models,
languages and algorithms. We refer the reader to the work by Brafman and
Domshlak (\citeyear{brafman2009preference}) for a
more detailed discussion on this metamodel, and by Baier and McIlraith
(\citeyear{baier2009planning}) for an overview of different preference
languages used in planning with preferences.

From the modeling point of view, in order to design a suitable language
capturing the user's preference model, the modeler should be provided with some
knowledge of the user's interest that affects the way she evaluates
plans (for instance, flight duration and ticket cost in a travel
planning scenario). Such knowledge in many cases, however, cannot be
completely specified. Our purpose therefore is
to present a bounded set of plans to the user in the hope that it will
increase the chance that she can find a desired plan. In the next
section, we formalize the quality measures for plan sets in two situations where
either no knowledge of the user's preferences or only part of them is given.

\section{Quality Measures for Plan Sets}
\label{sec:quality-measures}

\subsection{Syntactic Distance Measures for Unknown Preference Cases}
\noindent
We first consider the situation in which the user has some preferences
for solution plans, but the planner is not provided with any knowledge
of such preferences. It is therefore impossible for the planner to
assume any particular form of preference language representing the
hidden preference model. There are two issues that need to be considered
in formalizing a quality measure for plan sets:

\begin{itemize}

\item What are the elements of plans that can be involved in a quality
  measure?

\item How should a quality measure be defined using those elements?

\end{itemize}

For the first question, we observe that even though users are normally
interested in some \emph{high level} features of plans that are relevant
to them, many of those features can be considered as ``functions'' of \emph{base
level} elements of plans. For instance, the set of actions in the plan
determine the makespan of a (sequential) plan, and the sequence of
states when the plan executes gives the total reward of goals
achieved. We consider the following three types of base level features
of plans which could be used in defining quality measure, independently of the
domain semantics:

\begin{itemize}

\item \emph{Actions that are present in plans}, which define various
  high level features of the plans such as its makespan, execution cost etc. that
  are of interest to the user whose preference model could be represented with
  preference languages such as in PSP and PDDL2.1.
  
\item \emph{Sequence of states that the agent goes through, which captures
    the behaviors resulting from the execution of plans.} In many
  preference languages defined using high level features of plans such
  as the reward of goals collected (e.g., PSP), of the whole state
  (e.g., MDP), or the temporal relation between propositions occur in
  states (e.g. PDDL3, $\calP\calP$ \citep{son2006preference} and \emph{LPP}
  \citep{fritz2006decision}), the sequence of states can affect the
  quality of plan evaluated by the user.

\item \emph{The causal links representing how actions contribute to the
    goals being achieved, which measures the causal structures of
    plans.}\footnote{A causal link $a_1 \stackrel{p}{\rightarrow} a_2$
    records that a predicate is produced by $a_1$ and consumed by $a_2$.
  } These plan
  elements can affect the quality of plans with respect to the languages
  mentioned above, as the causal links capture both the actions appearing in
  a plan and the temporal relation between actions and variables.

\end{itemize}

A similar conceptual separation of features has also been
considered recently in the context of case-based planning by Serina
(\citeyear{serina2010kernel}), in which planning problems were assumed
to be well classified, in terms of costs to adapt plans of one problem
to solve another, in some \emph{unknown} high level feature space.
The similarity between problems in the space were implicitly defined
using kernel functions of their domain-independent graph
representations. In our situation, we aim to approximate quality of plan
sets on the space of features that the user is interested in using
distance between plans with respect to base level features of plans
mentioned above (see below).

Table~\ref{tab:plan-bases} gives the pros and cons of using the
different base level elements of plan. We note that if actions in the
plans are used in defining quality measure of plan sets, no additional
problem or domain theory information is needed. If plan behaviors are
used as base level elements, the representation of the plans that bring
about state transition becomes irrelevant since only the actual states
that an execution of the plan will take is considered. Hence, we can now
compare plans of different representations, e.g., four plans where the
first is a deterministic plan, the second is a contingent plan, the
third is a hierarchical plan and the fourth is a policy encoding
probabilistic behavior. If causal links are used, then the causal
proximity among actions is now considered rather than just physical
proximity in the plan.

\begin{table}
{
\begin{center}
\begin{tabular}{|l|l|l|} \hline
{\bf Basis}      & {\bf Pros}      & {\bf Cons}\\ \hline

 Actions         &   Does not require                & No problem information \\
                 &    problem information            &  is used \\ \hline 
 States          &   Not dependent on any specific   & Needs an execution \\
                 &    plan representation           &  simulator to identify states \\  \hline 
 Causal links   &   Considers causal proximity       & Requires domain theory \\
                 &    of state  transitions (action)  & \\
                 &    rather than positional          & \\
                 &    (physical) proximity & \\ \hline 

\end{tabular}
\caption{The pros and cons of different base level elements of plan.}
\label{tab:plan-bases}
\end{center}
}
\end{table}

Given those base level elements, the next question is how to define a
quality measure of plan sets using them. Recall that without any
knowledge about the user's preferences, there is no way for the planner
to assume any particular preference language, because of which the
motivation behind a choice of quality measure should come from the
hidden user's preference model. Given a Hasse diagram induced from the
user's preference model, a $k$-plan set that will be presented to the
user can be considered to be randomly selected from the diagram. The
probability of having one plan in the set classified in a class at the
optimal layer would increase when the individual plans are more likely
to be at different layers, and this chance in turn will increase if
they are less likely to be equally prefered by the user.\footnote{To see
  this, consider a diagram with $\calS_1=\{p_1,p_2\}$ at layer 0,
  $\calS_2=\{p_3\}$ and $\calS_3=\{p_4\}$ at layer 1, and
  $\calS_4=\{p_5\}$ at layer 2. Assuming that we randomly select a set
  of 2 plans. If those plans are known to be at the same layer, then the
  chance of having one plan at layer 0 is $\frac{1}{2}$. However, if
  they are forced to be at different layers, then the probability will
  be $\frac{3}{4}$.} On the other hand, the effect of base level
elements of a plan on high level features relevant to the user suggests
that \emph{plans similar with respect to base level features are more
  likely to be close to each other on the high level feature space
  determining user's preference model.}

In order to define a quality measure using base level features of plans,
we proceed with the following assumption: \emph{plans that are
  different from each other with respect to the base level features are
  less likely to be equally prefered by the user, in other words they
  are more likely to be at different layers of the Hasse diagram}. With
the purpose of increasing the chance of having a plan that the user
prefers, we propose the quality measure of plan sets as its \emph{diversity}
measure, defined using the distance between two plans in the set with
respect to a base level element. More formally, the quality measure
$\zeta: 2^{\calS} \rightarrow R$ of a
plan set $\calP$ can be defined as either the minimal, maximal, or
average distance between plans:

\begin{itemize}

\item Minimal distance:

  \begin{equation}
    \label{eq:min-quality-measure}
    \zeta_{min}(\calP) = \mymin\limits_{p,p' \in \calP}\delta(p,p')
  \end{equation}

\item Maximal distance:

  \begin{equation}
    \label{eq:max-quality-measure}
    \zeta_{max}(\calP) = \mymax\limits_{p,p' \in \calP}\delta(p,p')
  \end{equation}

\item Average distance:

  \begin{equation}
    \label{eq:avg-quality-measure}
    \zeta_{avg}(\calP) = {|\calP| \choose 2}^{-1} \times  \sum\limits_{p,p' \in \calP}\delta(p,p')
  \end{equation}

\end{itemize}

\noindent
where $\delta: \calS \times \calS \rightarrow \lbrack 0,1 \rbrack$ is
the distance measures between two plans.

\subsubsection{Distance measures between plans}
\label{sec:pairwise-distance}
\noindent
There are various choices on how to define the distance measure
$\delta(p,p')$ between two plans using plan actions, sequence of
states or causal links, and each way can have different impact on the
diversity of plan set on the Hasse diagram. In the following, we propose
distance measures in which a plan is considered as (i) a set of
actions and causal links, or (ii) sequence of states the agent goes
through, which could be used independently of plan representation (e.g.
total order, partial order plans).

\begin{itemize}

\item \textit{Plan as a set of actions or causal links}: given a
  plan $p$, let $A(p)$ and $C(p)$ be the set of actions or causal
  links of $p$. The distance between two plans $p$ and $p'$ can be
  defined as the ratio of the number of actions (causal links) that do not
  appear in both plans to the total number of actions (causal links)
  appearing in one of them:

\begin{equation}
\label{eq:action-distance}
\delta_A(p,p') = 1 - \frac{|A(p) \cap A(p')|}{|A(p) \cup A(p')|}
\end{equation}

\begin{equation}
\label{eq:clink-distance}
\delta_{CL}(p,p') = 1 - \frac{|C(p) \cap C(p')|}{|C(p) \cup C(p')|}
\end{equation}

\item \textit{Plan as a sequence of states}: given two sequence of
  states $(s_0,s_1,...,s_k)$ and $(s'_0, s'_1, ..., s'_{k'})$ resulting
  from executing two plans $p$ and $p'$, and assume that $k' \leq
  k$. Since the two sequence of states may have different length, there
  are various options in defining distance measure between $p$ and
  $p'$, and we consider here two simple options. In the first one, it
  can be defined as the average of the distances
  between state pairs $(s_i,s'_{i}) \ (0 \leq i \leq k')$, and each
  state $s_{k'+1}$,... $s_{k}$ is considered to contribute maximally
  (i.e. one unit) into the difference between two plans:

  \begin{equation}
    \label{eq:state-distance-1}
    \delta_S(p,p') = \frac{1}{k} \times \lbrack
    \sum_{i=1}^{k'}\Delta(s_i,s'_{i}) + k - k' \rbrack
  \end{equation}

  On the other hand, we can assume that the agent continues to stay at the goal state $s'_{k'}$ in
  the next $(k-k')$ time steps after executing $p'$, and the measure
  can be defined as follows:

  \begin{equation}
    \label{eq:state-distance-2}
    \delta_S(p,p') = \frac{1}{k} \times \lbrack
    \sum_{i=1}^{k'}\Delta(s_i,s'_{i}) + \sum_{i=k'+1}^{k}\Delta(s_i,s'_{k'}) \rbrack
  \end{equation}

\noindent
The distance measure $\Delta(s,s')$ between two states $s$, $s'$ used in
those two measures is defined as

\begin{equation}
\Delta(s,s') = 1 - \frac{s \cap s'}{s \cup s'}
\end{equation}

\end{itemize}

\begin{figure*}[t]
\centering
\epsfig{file=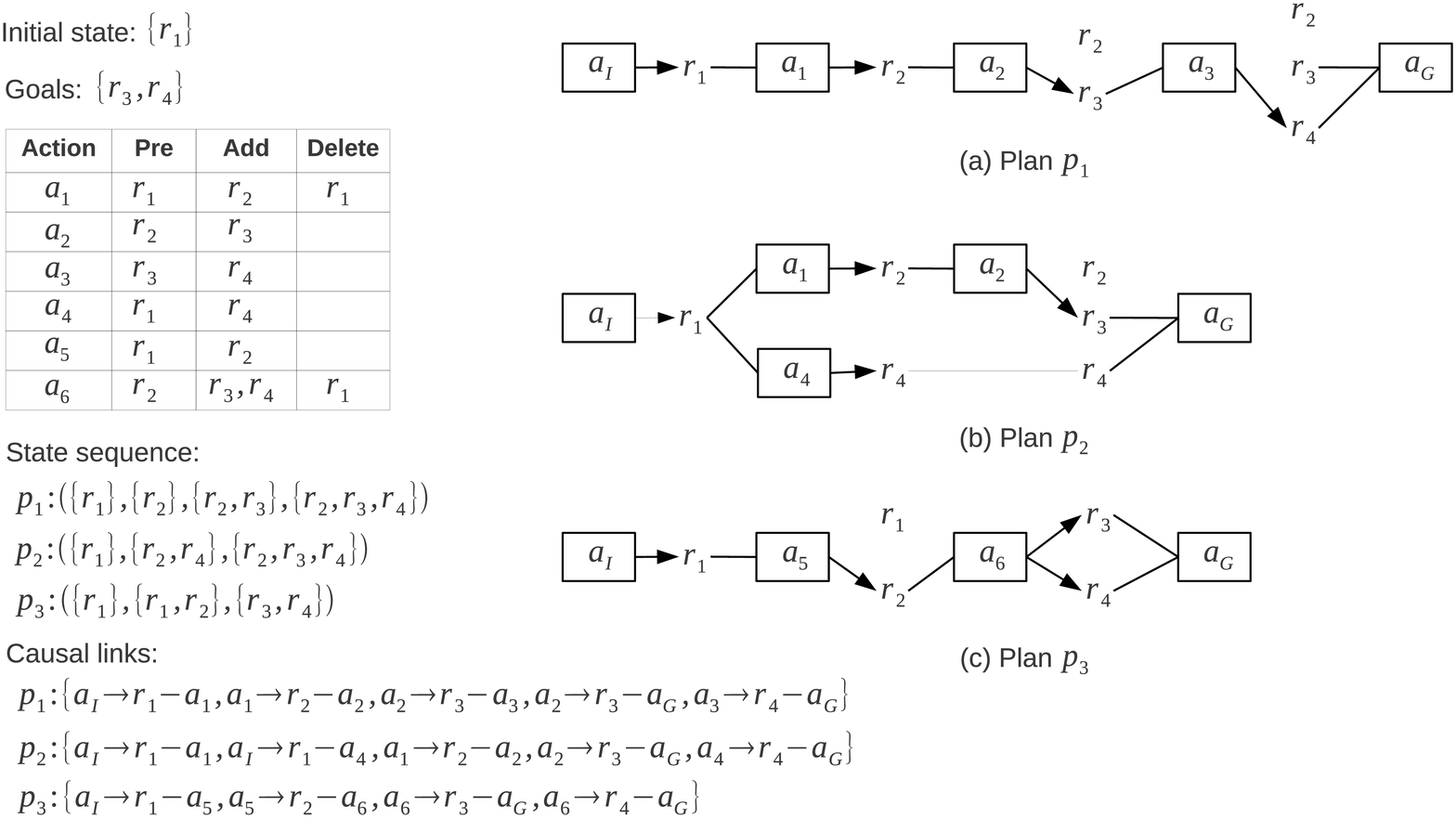,width=.98\linewidth}
\caption{Example illustrating plans with base-level elements. $a_I$ and
  $a_G$ denote dummy actions producing the initial state and consuming
  the goal propositions, respectively (see text for more details).}
\label{fig:distance-example}
\end{figure*}


\noindent{\bf Example:}
Figure~\ref{fig:distance-example} shows three plans $p_1$, $p_2$ and $p_3$ for a
planning problem where the initial state is $\{r_1\}$ and the goal
propositions are $\{r_3,r_4\}$. The specification of actions are shown
in the table. The action sets of the first two plans ($\{a_1,a_2,a_3\}$
and $\{a_1,a_2,a_4\}$) are quite similar ($\delta_A(p_1,p_2) = 0.5$), but
the causal links which involve $a_3$ ($a_2 \rightarrow r_3-a_3$, $a_3
\rightarrow r_4- a_G$) and $a_4$ ($a_I \rightarrow r_1-a_4$, $a_4
\rightarrow r_4-a_G$) make their difference more significant with respect to causal-link based
distance ($\delta_{CL}(p_1,p_2) = \frac{4}{7}$). 
Two other plans $p_1$ and $p_3$, on the other hand, are
very different in terms of action sets (and
therefore the sets of causal links): $\delta_A(p_1,p_3) = 1$, but
they are closer in term of state-based distance ($\frac{13}{18}$ as defined in
the equation~\ref{eq:state-distance-1}, and $0.5$ if defined in
the equation~\ref{eq:state-distance-2}).







\subsection{Integrated Preference Function (IPF) for Partial Preference Cases}

\noindent
We now discuss a quality measure for plan sets in the case when the
user's preference is partially expressed. In particular, we consider
scenarios in which the preference model can be represented by some
quantitative language with an incompletely specified value function of
high level features. As an example, the quality of plans in PDDL2.1
\citep{pddl2.1} and PDDL3 \citep{pddl3} are represented by a metric
function combining metric fluents and preference statements on state
trajectory with parameters representing their relative importance. While
providing a convenient way to represent preference models, such
parameterized value functions present an issue of obtaining reasonable
values for the relative importance of the features. A common approach to
model this type of incomplete knowledge is to consider those parameters
as a vector of random variables, whose values are assumed to be drawn
from a distribution. This is the representation that we will follow.

To measure the quality of plan sets, we propose the usage of
\emph{Integrated Preference Function} (IPF) \citep{ipf}, which has been
used to measure the quality of a solution set in a wide range of
multi-objective optimization problems. The IPF measure assumes that the
user's preference model can be represented by two factors: (1) a
probability distribution $h(\alpha)$ of parameter vector $\alpha$ such
that $\int_{\alpha} h(\alpha) \,d\alpha = 1$ (in the absence of any
special information about the distribution, $h(\alpha)$ can be assumed
to be uniform), and (2) a value function $V(p,\alpha): \cal{S}
\rightarrow \mathbb{R}$ combines different objective functions into a
single real-valued quality measure for plan $p$. This incomplete
specification of the value function represents a set of candidate
preference models, for each of which the user will select a different
plan, the one with the best value, from a given plan set $\calP
\subseteq \calS$.
The IPF value of solution set $\calP$ is defined as:

\begin{equation}
\label{eq:general-ipf}
IPF(\calP) = \int_{\alpha} h(\alpha) V(p_{\alpha},\alpha) \,d\alpha
\end{equation}

\noindent
with $p_{\alpha} = \argmin\limits_{ p \in \calP} {V(p,\alpha)}$ is the
best solution according to $V(p,\alpha)$ for each given $\alpha$ value.
Let $p^{-1}_{\alpha}$ be its inverse function specifying a range of
$\alpha$ values for which $p$ is an optimal solution according to
$V(p,\alpha)$. As $p_{\alpha}$ is piecewise constant, the $IPF(\calP)$
value can be computed as:

\begin{equation}
\label{eq:sum-ipf}
IPF(\calP) = \sum_{p \in \calP} \left  [ \int_{\alpha \in p^{-1}_{\alpha}}
  h(\alpha) V(p,\alpha) \, d\alpha \right ].
\end{equation}

Let $\calP^* = \{p \in \calP:\ p^{-1}_{\alpha} \neq
\emptyset\}$ then we have:

\begin{equation}
\label{eq:sum-ipf-1}
IPF(\calP) = IPF(\calP^*) = \sum_{p \in \calP^*} \left  [ \int_{\alpha \in
    p^{-1}_{\alpha}} h(\alpha) V(p,\alpha) \, d\alpha \right ].
\end{equation}

Since $\calP^*$ is the set of plans that are optimal for some specific
parameter vector, $IPF(\calP)$ now can be interpreted as the
expected value that the user can get by selecting the best plan in
$\calP$. Therefore, the set $\calP^*$ of solutions (known as \emph{lower convex hull} of
$\calP$) with the minimal IPF value is most likely to contain the
desired solutions that the user wants and in essense a good
representative of the plan set $\calP$.

While our work is applicable to more general planning scenarios, to make
our discussion on generating plan sets concrete, we will concentrate on
metric temporal planning where each action $a \in A$ has a duration
$d_a$ and execution cost $c_a$. The planner needs to find a plan $p =
\{a_1 \ldots a_n\}$, which is a sequence of actions that is executable
and achieves all goals. The two most common plan quality measures are:
\emph{makespan}, which is the total execution time of $p$; and
\emph{plan cost}, which is the total execution cost of all actions in
$p$---both of them are high level features that can be affected by the actions in the
plan. In most real-world applications, these two
criteria compete with each other: shorter plans usually have higher cost
and vice versa. We use the following assumptions:

\begin{itemize}

\item The desired objective function involves minimizing both
  components: $time(p)$ measures the makespan of the plan $p$ and
  $cost(p)$ measures its execution cost.

\item The quality of a plan $p$ is a convex combination: $V(p,w) = w
  \times time(p) + (1-w) \times cost(p)$, where weight $w \in [0,1]$
  represents the trade-off between the two competing objective
  functions.

\item The belief distribution of $w$ over the range $[0,1]$ is known. If
  the user does not provide any information or we have not learnt
  anything about the preference on the trade-off between \emph{time} and
  \emph{cost} of the plan, then the planner can assume a uniform
  distribution (and improve it later using techniques such as preference
  elicitation).

\end{itemize}

Given that the exact value of $w$ is unknown, our purpose is to find a
bounded representative set of non-dominated plans\footnote{A plan $p_1$
  is dominated by $p_2$ if $time(p_1) \geq time(p_2)$ and $cost(p_1)
  \geq cost(p_2)$ and at least one of the inequalities is strict.}
minimizing the expected value of $V(p,w)$ with regard to the given
distribution of $w$ over $[0,1]$.

\begin{figure}[t]
\centering
\epsfig{file=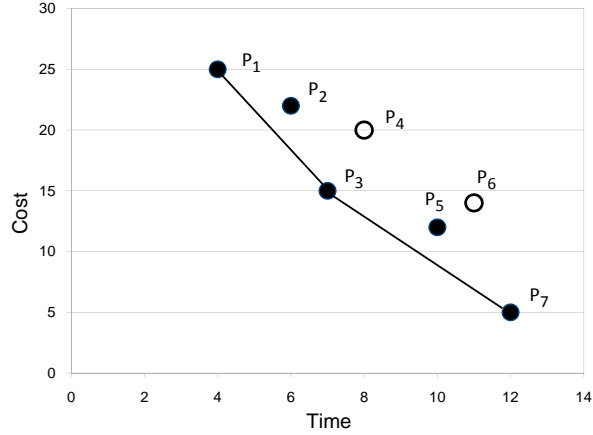,width=3in}
\caption{Solid dots represents plans in the pareto set ($p_1, p_2, p_3,
  p_5, p_7$). Connected dots represent plans in the lower convex hull
  ($p_1, p_3, p_7$) giving optimal ICP value for any distribution on
  trade-off between \emph{cost} and \emph{time}.}
\label{fig:icp_example}
\end{figure}

\noindent{\bf Example:} Figure~\ref{fig:icp_example} shows our running
example in which there are a total of 7 plans with their $time(p)$ and
$cost(p)$ values as follows: $p_1 = \{4,25\}$, $p_2 = \{6,22\}$, $p_3
= \{7,15\}$, $p_4 = \{8,20\}$, $p_5 = \{10,12\}$, $p_6 = \{11,14\}$,
and $p_7 = \{12,5\}$. Among these 7 plans, 5 of them belong to a
pareto optimal set of
non-dominated plans:
$\calP_{p} = \{p_1, p_2, p_3, p_5, p_7\}$. The other two plans are
dominated by some plans in $\calP_{p}$: $p_4$ is dominated by $p_3$ and
$p_6$ is dominated by $p_5$. Plans in $\calP_p$ are depicted in
solid dots, and the set of plans $\calP^* = \{p_1,p_3,p_7\}$ that are
optimal for some specific value of $w$ is highlighted by connected dots.

\medskip
\noindent{\bf IPF for Metric Temporal Planning:}
The user preference model in our target domain of temporal planning
is represented by a convex combination of the \emph{time} and \emph{cost} quality measures,
and the IPF measure now is called \emph{Integrated Convex Preference} (ICP).
Given a set of plans $\calP^*$, let $t_p = time(p)$ and $c_p = cost(p)$
be the makespan and total execution cost of plan $p \in \calP^*$, the
ICP value of $\calP^*$ with regard to the objective function $V(p,w) = w
\times t_p + (1-w) \times c_p$ and the parameter vector $\alpha=(w,1-w)$
($w \in [0,1]$) is defined as:

\begin{equation}
  \label{eq:convex-combination}
  ICP(\calP^*) = \sum_{i=1}^{k}\int_{w_{i-1}}^{w_i} h(w) (w \times t_{p_i} + (1-w) \times c_{p_i})dw
\end{equation}

\noindent where $w_0 = 0$, $w_k = 1$ and $p_i = \argmin\limits_{p \in
  \calP^*} V(p,w) \ \forall w \in [w_{i-1},w_i]$. In other words, we
divide $[0,1]$ into non-overlapping regions such that in each region
$(w_{i-1},w_i)$ there is a single solution $p_i \in \calP^*$ that has
better $V(p_i,w)$ value than all other solutions in $\calP^*$.

We select the IPF/ICP measure to evaluate our solution set due to its
several nice properties:

\begin{itemize}
\item If $\calP_1, \calP_2 \subseteq \calS$ and $ICP(\calP_1) <
  ICP(\calP_2)$ then $\calP_1$ is \emph{probabilistically} better than
  $\calP_2$ in the sense that for any given $w$, let $p_1 =
  \argmin\limits_{p \in \calP_1} V(p,w)$ and $p_2 = \argmin\limits_{p
    \in \calP_2} V(p,w)$, then the probability of $V(p_1,w) < V(p_2,w)$
  is higher than the probability of $V(p_1,w) > V(p_2,w)$.

\item If $\calP_1$ is \emph{obviously} better than $\calP_2$, then the
  ICP measure agrees with the assertion. More formally: if $\forall p_2
  \in \calP_2, \exists p_1 \in \calP_1$ such that $p_2$ is dominated by
  $p_1$, then $ICP(\calP_1) < ICP(\calP_2)$.
\end{itemize}

Empirically, extensive results on scheduling problems in
\cite{fowler2005evaluating} have shown that ICP measure \emph{``evaluates the
solution quality of approximation robustly (i.e., similar to visual
comparison results) while other alternative measures can misjudge the
solution quality''}.

In the next two sections~\ref{sec:planning-no-preference} and
\ref{sec:planning-partial-preference}, we investigate the problem of
generating high quality plan sets for two cases mentioned: when no knowledge
about the user's preferences is given, and when part of its is given as input to
the planner.

\section{Generating Diverse Plan Set in the Absence of Preference Knowledge}
\label{sec:planning-no-preference}
\noindent
In this section, we describe approaches to searching for a set of
diverse plans with respect to a measure defined with base level elements
of plans as discussed in the previous section. In particular, we
consider the quality measure of plan set as the minimal pair-wise
distance between any two plans, and generate a set of plans containing
$k$ plans with the quality of at least a predefined threshold $d$. As
discussed earlier, by diversifying the set of plans on the space of base
level features, it is likely that plans in the set would cover a wide
range of space of unknown high level features, increasing the
possibility that the user can select a plan close to the one that she
prefers. The problem is formally defined as follows:

\begin{quote} 

  {\em d}DISTANT{\em k}SET : Find $\calP$ with $\calP \subseteq \calS$,
  $\mid \calP \mid$ = $k$ and $\zeta(\calP) = \mymin\limits_{p,q \in
    \calP} \delta(p,q) \geq d$

\end{quote}

\noindent
where any distance measure between two plans formalized in
Section~\ref{sec:pairwise-distance} can be used to implement
$\delta(p,p')$.

We now consider two representative state-of-the-art planning approaches
in generating diverse plan sets. The first one is {\sc gp-csp}
\citep{gp-csp} representing constraint-based planning approaches, and
the second one is {\sc lpg} \citep{lpg:JAIR03} that uses an efficient
local-search based approach. We use {\sc gp-csp} to comparing the
relation between different distance measures in diversifying plan sets.
On the other hand, with {\sc lpg} we stick to the action-based distance
measure, which is shown experimentally to be the most difficult measure
to enforce diversity (see below), and investigate the scalability of
heuristic approaches in generating diverse plans.

\subsection{Finding Diverse Plan Set with GP-CSP}
\label{sec:gpcsp-background}

\noindent
The {\sc gp-csp} planner \citep{gp-csp} converts Graphplan's planning
graph into a CSP encoding, and solves it using a standard CSP solver.
The solution of the encoding represents a valid plan for the original
planning problem. In the encoding, the CSP variables correspond to the
predicates that have to be achieved at different levels in the planning
graph (different planning steps) and their possible values are the
actions that can support the predicates. For each CSP variable
representing a predicate $p$, there are two special values: i) $\bot$:
indicates that a predicate is not supported by any action and is
\emph{false} at a particular level/planning-step; ii) ``noop'':
indicates that the predicate is true at a given level $i$ because it was
made true at some previous level $j < i$ and no other action deletes $p$
between $j$ and $i$. Constraints encode the relations between predicates
and actions: 1) mutual exclusion relations between predicates and
actions; and 2) the causal relationships between actions and their
preconditions.

\subsubsection{Adapting GP-CSP to Different Distance Metrics}
\label{sec:gpcsp-adapting}

\noindent
When the above planning encoding is solved by any standard CSP solver,
it will return a solution containing $\langle${\em var, value}$\rangle$
of the form $\{\langle x_1,y_1 \rangle,... \langle x_n,y_n \rangle\}$.
The collection of $x_i$ where $y_i \ne \bot$ represents the facts that
are made true at different time steps (plan trajectory) and can be used
as a basis for the \emph{state-based} distance measure\footnote{We
  implement the state-based distance between plans as defined in
  equation~\ref{eq:state-distance-1}.}; the set of $(y_i \ne \bot)
\wedge (y_i \ne noop)$ represents the set of actions in the plan and can
be used for \emph{action-based} distance measure; lastly, the
assignments $\langle x_i,y_i \rangle$ themselves represent the causal
relations and can be used for the \emph{causal-based} distance measure.

However, there are some technical difficulties we need to overcome before a
specific distance measure between plans can be computed. First, the same
action can be represented by different values in the domains of
different variables. Consider a simple example in which there are two
facts $p$ and $q$, both supported by two actions $a_1$ and $a_2$. When
setting up the CSP encoding, we assume that the CSP variables $x_1$ and
$x_2$ are used to represent $p$ and $q$.  The domains for $x_1$ and
$x_2$ are $\{v_{11},v_{12}\}$ and $\{v_{21},v_{22}\}$, both representing
the two actions $\{a_1,a_2\}$ (in that order). The assignments
$\{\langle x_1,v_{11} \rangle, \langle x_2, v_{21} \rangle\}$ and
$\{\langle x_1,v_{12} \rangle, \langle x_2, v_{22} \rangle\}$ have a
distance of 2 in traditional CSP because different values are assigned
for each variable $x_1$ and $x_2$. However, they both represent the same
action set $\{a_1,a_2\}$ and thus lead to the plan distance of 0 if we
use the action-based distance in our plan comparison. Therefore, we
first need to translate the set of values in all assignments back to the
set of action instances before doing comparison using action-based
distance.  The second complication arises for the causal-based distance.
A causal link $a_1 \stackrel{p}{\rightarrow} a_2$ between two actions
$a_1$ and $a_2$ indicates that $a_1$ supports the precondition $p$ of
$a_2$. However, the CSP assignment $\langle p, a_1 \rangle$ only
provides the first half of each causal-link. To complete the
causal-link, we need to look at the values of other CSP assignments to
identify action $a_2$ that occurs at the later level in the planning
graph and has $p$ as its precondition. Note that there may be multiple
``valid'' sets of causal-links for a plan, and in the implementation
we simply select causal-links based on the CSP assignments.

\subsubsection{Making GP-CSP Return a Set of Plans}
\label{sec:gpcsp-par-gre}

\noindent
To make {\sc gp-csp} return a set of plans satisfying the {\em
  d}DISTANT{\em k}SET constraint using one of the three distance measures, we
add ``global'' constraints to each original encoding to enforce {\em
  d}-diversity between every pair of solutions. When each global
constraint is called upon by the normal forward checking and
arc-consistency checking procedures inside the default solver to check
if the distance between two plans is over a predefined value $d$, we
first map each set of assignments to an actual set of actions
(action-based), predicates that are true at different plan-steps
(state-based) or causal-links (causal-based) using the method discussed
in the previous section. This process is done by mapping all $\langle
var ,value \rangle$ CSP assignments into action sets using a call to the
planning graph, which is outside of the CSP solver, but works closely
with the general purpose CSP solver in {\sc gp-csp}. The comparison is
then done within the implementation of the global constraint to decide
if two solutions are diverse enough.

We investigate two different ways to use the global constraints: 

\begin{enumerate}

\item The {\em parallel } strategy to return the set of $k$ plans all at
  once. In this approach, we create one encoding that contains $k$
  identical copies of each original planning encoding created using {\sc
    gp-csp} planner. The $k$ copies are connected together using
  $k(k-1)/2$ pair-wise global constraints. Each global constraint
  between the $i^{th}$ and $j^{th}$ copies ensures that two plans
  represented by the solutions of those two copies will be at least $d$
  distant from each other. If each copy has $n$ variables, then this
  constraint involves $2n$ variables.

\item The {\em greedy } strategy to return plans one after another. In
  this approach, the $k$ copies are not setup in parallel up-front, but
  sequentially. We add to the $i^{th}$ copy one global constraint to
  enforce that the solution of the $i^{th}$ copy should be $d$-diverse
  from any of the earlier $i-1$ solutions.  The advantage of the greedy
  approach is that each CSP encoding is significantly smaller in terms
  of the number of variables ($n$ vs. $k*n$), smaller in terms of the
  number of global constraints (1 vs. $k(k-1)/2$), and each global
  constraint also contains lesser number of variables ($n$ vs.
  $2*n$).\footnote{However, each constraint is more complicated because
    it encodes {\em (i-1)} previously found solutions.} Thus, each
  encoding in the greedy approach is easier to solve.  However, because
  each solution depends on all previously found solutions, the encoding
  can be unsolvable if the previously found solutions comprise a bad
  initial solution set.

\end{enumerate}

%

\subsubsection{Empirical Evaluation}
\label{sec:gpcsp-res}

\noindent
We implemented the parallel and greedy approaches discussed earlier for
the three distance measures and tested them with the benchmark set of
\emph{Logistics} problems provided with the Blackbox
planner \citep{blackbox}.  All experiments were run on a Linux Pentium 4,
3Ghz machine with 512MB RAM. For each problem\footnote{log-easy=prob1,
  rocket-a=prob2, log-a = prob3, log-b = prob4, log-c=prob5,
  log-d=prob6.}, we test with different $d$ values ranging from 0.01
(1\%) to 0.95 (95\%)\footnote{Increments of 0.01 from 0.01 to 0.1 and of
  0.05 thereafter.} and $k$ increases from 2 to $n$ where $n$ is the
maximum value for which {\sc gp-csp} can still find solutions within
plan horizon. The horizon (parallel plan steps) limit is 30.

We found that the greedy approach outperformed the parallel approach and
solved significantly higher number of problems. Therefore, we focus on
the greedy approach hereafter. For each combination of $d$, $k$, and a
given distance measure, we record the solving time and output the
average/min/max pairwise distances of the solution sets.

\medskip
\noindent{\bf Baseline Comparison:}
As a baseline comparison, we have also implemented a \emph{randomized}
approach. In this approach, we do not use global constraints but use
random value ordering in the CSP solver to generate $k$ different
solutions without enforcing them to be pairwise $d$-distance apart. For
each distance $d$, we continue running the random algorithm until we
find $k_{max}$ solutions where $k_{max}$ is the maximum value of $k$
that we can solve for the greedy approach for that particular $d$ value.
In general, we want to compare with our approach of using global
constraint to see if the random approach can effectively generate
diverse set of solutions by looking at: (1) the average time to find a
solution in the solution set; and (2) the maximum/average pairwise
distances between $k \ge 2$ randomly generated solutions.

\begin{table}
{\small
\begin{center}
\begin{tabular} {| c || c | c | c | c | c | c |}
\hline & {\bf Prob1} & {\bf Prob2} & {\bf Prob3} & {\bf Prob4} & {\bf Prob5} & {\bf Prob6} \\
\hline \hline $\delta_a$ & 0.087 & 7.648 & 1.021 & 6.144 & 8.083 &
178.633 \\
\hline $\delta_s$ & 0.077 & 9.354 & 1.845 & 6.312 & 8.667 & 232.475 \\
\hline $\delta_c$ & 0.190 & 6.542 & 1.063 & 6.314 & 8.437 & 209.287 \\
\hline \hline {\em Random} & 0.327 & 15.480 & 8.982 & 88.040 &
379.182 & 6105.510 \\
\hline
\end{tabular}
\caption {Average solving time (in seconds) to find a plan 
using greedy (first 3 rows) and by random (last row) approaches}
\label{table:time_per_sol}
\end{center}
 }
 \mvp
\end{table}

\begin{table}
{\small
\begin{center}
\begin{tabular}{| c || c | c | c | c | c | c |}
\hline & {\bf Prob1} & {\bf Prob2} & {\bf Prob3} & {\bf Prob4} & {\bf Prob5} & {\bf Prob6} \\
\hline \hline $\delta_a$ & 0.041/0.35 & 0.067/0.65 & 0.067/0.25
& 0.131/0.1* & 0.126/0.15 & 0.128/0.2 \\
\hline $\delta_s$ & 0.035/0.4 & 0.05/0.8 & 0.096/0.5 & 0.147/0.4 & 0.140/0.5 & 0.101/0.5 \\
\hline $\delta_c$ & 0.158/0.8 & 0.136/0.95 & 0.256/0.55 &
0.459/0.15* & 0.346/0.3* & 0.349/0.45 \\
\hline
\end{tabular}
\caption{Comparison of the diversity in the plan sets returned by
the random and greedy approaches. Cases where random approach is better
than greedy approach are marked with (*).} 
\label{table:distance_random}
\end{center}
}
\mvp
\end{table}

Table~\ref{table:time_per_sol} shows the comparison of average solving
time to find one solution in the greedy and random approaches. The
results show that on an average, the random approach takes significantly
more time to find a single solution, regardless of the distance measure
used by the greedy approach. To assess the diversity in the solution
sets, Table~\ref{table:distance_random} shows the comparison of: (1) the
average pairwise minimum distance between the solutions in sets returned
by the random approach; and (2) the maximum $d$ for which the greedy
approach still can find a set of diverse plans. The comparisons are done
for all three distance measures.  For example, the first cell $(0.041/0.35)$ in Table~\ref{table:distance_random}, implies that the minimum
pairwise distance averaged for all solvable $k \ge 2$ using the random
approach is $d = 0.041$ while it is $0.35$ (i.e. 8x more diverse) for
the greedy approach using the $\delta_a$ distance measure. Except for 3
cases, using global constraints to enforce minimum pairwise distance
between solutions helps {\sc gp-csp} return significantly more diverse
set of solutions. On average, the greedy approach returns 4.25x,
7.31x, and 2.79x more diverse solutions than the random approach for
$\delta_a$, $\delta_s$ and $\delta_c$, respectively.

\medskip
\noindent{\bf Analysis of the different distance-bases:} Overall, we
were able to solve 1264 $(d,k)$ combinations for three distance measures
$\delta_a, \delta_s, \delta_c$ using the greedy approach. We were
particularly interested in investigating the following issues:
\begin{itemize}
\item {\bf H1: Computational efficiency} - Is it easy or difficult to find a set of diverse
  solutions using different distance measures? Thus, (1) for the same
  $d$ and $k$ values, which distance measure is more difficult (time
  consuming) to solve; and (2) given an encoding horizon limit, how high
  is the value of $d$ and $k$ for which we can still find a set of solutions
  for a given problem using different distance measures.
\item {\bf H2: Solution diversity} - What, if any, is the correlation/sensitivity between
  different distance measures? Thus, how comparative diversity of
  solutions is when using different distance measures.


\end{itemize}

\begin{table} {\small
\begin{center}
\begin{tabular}{| l || l | l | l | l | l | l |} \hline {\bf d} & {\bf Prob1} & {\bf Prob2} & {\bf Prob3} & {\bf Prob4} & {\bf Prob5} & {\bf Prob6} \\
\hline \hline

      0.01 & 11,5, {\bf 28} & 8,{\bf 18},12 & 9,8,{\bf 18} & 3,4,{\bf 5} & 4,6,{\bf 8} & {\bf 8},7,7 \\

 \hline 0.03 & 6,3,{\bf 24} & 8,{\bf 13},9 & 7,7,{\bf 12} & 2,{\bf 4},3 & 4,{\bf 6},{\bf 6} & 4,{\bf 7},6 \\

\hline 0.05 & 5,3,{\bf 18} & 6,{\bf 11},9 & 5,7,{\bf 10} & 2,{\bf 4},3 & 4,{\bf 6},5 & 3,{\bf 7},5 \\

\hline 0.07 & 2,3,{\bf 14} & 6,{\bf 10},8 & 4,{\bf 7},6 & 2,{\bf 4},2 & 4,{\bf 6},5 & 3,{\bf 7},5 \\


\hline 0.09 & 2,3,{\bf 14} & 6,{\bf 9},6 & 3,{\bf 6},{\bf 6} & 2,{\bf 4},2 & 3,{\bf 6},4 & 3,{\bf 7},4 \\ 

\hline 0.1 & 2,3,{\bf 10} & 6,{\bf 9},6 & 3,{\bf 6},{\bf 6} & 2,{\bf 4},2 & 2,{\bf 6},4 & 3,{\bf 7},4 \\ 

\hline 

       0.2 & 2,3,{\bf 5} & 5,{\bf 9},6 & 2,{\bf 6},{\bf 6} & 1,{\bf 3},1 & 1,{\bf 5},2 & 2,{\bf 5},3 \\ 

\hline 

       0.3 & 2,2,{\bf 3} & 4,{\bf 7},5 & 1,{\bf 4},{\bf 4} & 1,{\bf 2},1 & 1,{\bf 3},2 & 1,{\bf 3},{\bf 3} \\ 

\hline 

       0.4 & 1,2,{\bf 3} & 3,{\bf 6},5 & 1,{\bf 3},{\bf 3} & 1,{\bf 2},1 & 1,{\bf 2},1 & 1,2,{\bf 3} \\ 

\hline 

       0.5 & 1,1,{\bf 3} & 2,4,{\bf 5} & 1,{\bf 2},{\bf 2} & - & 1,{\bf 2},1 & 1,{\bf 2},1 \\ 

\hline 

       0.6 & 1,1,{\bf 2} & 2,3,{\bf 4} & - & - & - & - \\ 

\hline 

       0.7 & 1,1,{\bf 2} & 1,{\bf 2},{\bf 2} & - & - & - & - \\ 

\hline 

       0.8 & 1,1,{\bf 2} & 1,{\bf 2},{\bf 2} & - & - & - & - \\ 

\hline 

       0.9 & - & 1,1,{\bf 2} & - & - & - & - \\ 

\hline 

\end{tabular}
\caption{For each given $d$ value, each cell shows the largest solvable
  $k$ for each of the three distance measures $\delta_a$, $\delta_s$,
  and $\delta_c$ (in this order). The maximum values in cells are in
  bold.} \mvp \mvp
\label{table:max-d-k}
\end{center} }

\end{table} 

Regarding \emph{H1}, Table~\ref{table:max-d-k} shows the highest
solvable $k$ value for each distance $d$ and base $\delta_a$,
$\delta_s$, and $\delta_c$. For a given $(d,k)$ pair, enforcing
$\delta_a$ appears to be the most difficult, then $\delta_s$, and
$\delta_c$ is the easiest.  {\sc gp-csp} is able to solve 237, 462, and
565 combinations of $(d,k)$ respectively for $\delta_a$, $\delta_s$ and
$\delta_c$.  {\sc gp-csp} solves {\em d}DISTANT{\em k}SET problems more
easily with $\delta_s$ and $\delta_c$ than with $\delta_a$ due to the
fact that solutions with different action sets (diverse with regard to
$\delta_a$) will likely cause different trajectories and causal
structures (diverse with regard to $\delta_s$ and $\delta_c$).  Between
$\delta_s$ and $\delta_c$, $\delta_c$ solves more problems for easier
instances (Problems 1-3) but less for the harder instances, as shown in
Table~\ref{table:max-d-k}. We conjecture that for solutions with more
actions (i.e. in bigger problems) there are more causal dependencies
between actions and thus it is harder to reorder actions to create a
different causal-structure.

For running time comparisons, among 216 combinations of $(d,k)$ that
were solved by all three distance measures, {\sc gp-csp} takes the least
amount of time for $\delta_a$ in 84 combinations, for $\delta_s$ in 70
combinations and in 62 for $\delta_c$. The first three lines of
Table~\ref{table:time_per_sol} show the average time to find one
solution in $d$-diverse $k$-set for each problem using $\delta_a$,
$\delta_s$ and $\delta_c$ (which we call $t_a$, $t_s$ and $t_c$
respectively). In general, $t_a$ is the smallest and $t_s > t_c$ in most
problems. Thus, while it is harder to enforce $\delta_a$ than $\delta_s$
and $\delta_c$ (as indicated in Table~\ref{table:max-d-k}), when the
encodings for all three distances can be solved for a given $(d,k)$,
then $\delta_a$ takes less time to search for one plan in the diverse
plan set; this can be due to tighter constraints
(more pruning power for the global constraints) and simpler global
constraint setting.




\begin{table} {\small
    \begin{center} \begin{tabular}{| l || l | l | l |} \hline &
        $\delta_a$ & $\delta_s$ & $\delta_c$ \\ \hline \hline $\delta_a$
        & - & 1.262 & 1.985 \\ \hline $\delta_s$ & 0.485 & - & 0.883 \\
        \hline $\delta_c$ & 0.461 & 0.938 & - \\ \hline

\end{tabular}
\caption{Cross-validation of distance measures $\delta_a$, $\delta_s$,
  and $\delta_c$.}  \mvp \mvp
\label{table:cross-val}
\end{center} }

\end{table} 

To test \emph{H2}, in Table~\ref{table:cross-val}, we show the
cross-comparison between different distance measures $\delta_a$,
$\delta_s$, and $\delta_c$. In this table, cell $\langle${\em row,
  column}$\rangle = \langle \delta', \delta'' \rangle$ indicates that
over all combinations of $(d,k)$ solved for distance $\delta'$, the
average value $d''/d'$ where $d''$ and $d'$ are distance measured
according to $\delta''$ and $\delta'$, respectively ($d' \geq d$). For
example, $\langle \delta_s, \delta_a \rangle = 0.485$ means that over
462 combinations of $(d,k)$ solvable for $\delta_s$, for each $d$, the
average distance between $k$ solutions measured by $\delta_a$ is $0.485
* d_s$. The results indicate that when we enforce $d$ for $\delta_a$, we
will likely find even more diverse solution sets according to $\delta_s$
($1.26*d_a$) and $\delta_c$ ($1.98*d_a$).  However, when we enforce $d$
for either $\delta_s$ or $\delta_c$, we are not likely to find a more
diverse set of solutions measured by the other two distance measures.
Nevertheless, enforcing $d$ using $\delta_c$ will likely give comparable
diverse degree $d$ for $\delta_s$ ($0.94*d_c$) and vice versa. We also
observe that $d_s$ is highly dependent on the difference between the
parallel lengths of plans in the set. The distance $d_s$ seems to be the smallest
(i.e. $d_s < d_a < d_c$) when all $k$ plans have the same/similar number
of time steps. This is consistent with the fact that $\delta_a$ and
$\delta_c$ do not depend on the steps in the plan execution trajectory
while $\delta_s$ does.

\subsection{Finding Diverse Plan Set with LPG}
\label{sec:lpg-background}

\noindent
In this section, we consider the problem of generating diverse set of
plans using another planning approach, in particular the {\sc lpg} planner
which is able to scale up to bigger problems, compared to {\sc gp-csp}. We focus on the action-based distance
measure between plans, which has been shown in the previous section to be the most difficult to
enforce diversity. \lpg\ is a local-search-based planner, that incrementally modifies a
partial plan in a search for a plan that contains no flaws
\citep{lpg:JAIR03}. The behavior of \lpg\ is controlled by an evaluation
function that is used to select between different plan candidates in a
neighborhood generated for local search.  At each search step, the
elements in the search neighborhood of the current partial plan $\pi$
are the alternative possible plans repairing a selected flaw in $\pi$.
The elements of the neighborhood are evaluated according to an {\em
  action evaluation function} $E$ \citep{lpg:JAIR03}. This function is
used to estimate the cost of either adding or of removing an action node
$a$ in the partial plan $p$ being generated.

\subsubsection{Revised Evaluation Function for Diverse Plans}
\label{sec:lpg-changes-1}

\noindent
In order to manage \ddkset\ problems, the function $E$ has been extended
to include an additional evaluation term that has the purpose of
penalizing the insertion and removal of actions that {\em decrease} the
distance of the current partial plan $p$ under adaptation from a
reference plan $p_0$.
In general, $E$ consists of four weighted terms, evaluating four aspects
of the quality of the current plan that are affected by the addition
($E(a)^i$) or removal ($E(a)^r$) of $a$

\vspace{-0.2cm}
{\small \begin{multline*}
E(a)^i = \alpha_E \cdot Execution\_cost(a)^i + \alpha_T \cdot Temporal\_cost(a)^i  +\\ 
+ \alpha_S \cdot Search\_cost(a)^i + \alpha_D \cdot | (p_0 - p) \cap p_{\mathsf R}^i |
\end{multline*}
\vspace{-0.6cm}
\begin{multline*}
E(a)^r = \alpha_E \cdot Execution\_cost(a)^r + \alpha_T \cdot Temporal\_cost(a)^r  +\\ 
+ \alpha_S \cdot Search\_cost(a)^r + \alpha_D \cdot  | (p_0 - p - a ) \cap 
p_{\mathsf R}^r |.
\end{multline*}}

The first three terms of the two forms of $E$ are unchanged from the
standard behavior of \lpg. The fourth term, used only for computing diverse plans, is the new term
estimating how the proposed plan modification will affect the distance
from the reference plan $p_0$.  Each cost term in $E$ is computed
using a relaxed temporal plan $p_R$ \citep{lpg:JAIR03}.


The $p_R$ plans are computed by an algorithm, called {\small
  \sf RelaxedPlan}, formally described and illustrated in
\cite{lpg:JAIR03}. We have slightly modified this algorithm to penalize the
selection of actions decreasing the plan distance from the reference
plan. The specific change to {\small \sf RelaxedPlan} for computing
diverse plans is very similar to the change described in \citep{fox2006plan}, and it concerns the heuristic function for selecting
the actions for achieving the subgoals in the relaxed plans. In the
modified function for {\small \sf RelaxedPlan}, we have an extra 0/1
term that penalizes an action $b$ for $p_R$ if its addition
decreases the distance of $p + p_R$ from $p_0$ (in the
plan repair context investigated in \citep{fox2006plan}, $b$ is penalized
if its addition {\em increases} such a distance).

The last term of the modified evaluation function $E$ is a measure of
the decrease in plan distance caused by adding or removing $a$: $|
(p_0 - p) \cap p_R^i |$ or $| (p_0 - p - a ) \cap
p_R^r |$, where $p_R^i$ contains the new action
$a$.  The $\alpha$-coefficients of the $E$-terms are used to weigh their
relative importance.\footnote{These coefficients are also normalized to
  a value in $[0,1]$ using the method described in \cite{lpg:JAIR03}.  }
The values of the first 3 terms are automatically derived from the
expression defining the plan metric for the problem \citep{lpg:JAIR03}.
The coefficient for the fourth new term of $E$ ($\alpha_D$) is
automatically set during search to a value proportional to $d/
\delta_a(p, p_0)$, where $p$ is the current partial plan under
construction. The general idea is to dynamically increase the value of
$\alpha_D$ according to the number of plans $n$ that have been generated
so far: if $n$ is much higher than $k$, the search process consists of
finding many solutions with not enough diversification, and hence the
importance of the last $E$-term should increase.

\subsubsection{Making LPG Return a Set of Plans}
\label{sec:lpg-changes-2}

\noindent
In order to compute a set of $k$ $d$-distant plans solving a
\ddkset\ problem, we run the \lpg\ search multiple times, until the
problem is solved, with the following two additional changes to the
standard version of \lpg: (i) the preprocessing phase computing mutex
relations and other reachability information exploited during the
relaxed plan construction is done only once for all runs; (ii) we
maintain an incremental set of valid plans, and we dynamically select
one of them as the reference plan $p_0$ for the next search.
Concerning (ii), let $\calP=\{p_1,...,p_n\}$ be the set of $n$ valid plans
that have been computed so far, and {\it CPlans}$(p_i)$ the subset of
$\calP$ containing all plans that have a distance greater than or equal to
$d$ from a reference plan $p_i \in P$.

The reference plan $p_0$ used in the modified heuristic function $E$
is a plan $p_{max} \in \calP$ which has a maximal set of diverse plans in
$\calP$, i.e.,

\begin{equation}
p_{max}=ARGMAX_{\{p_i \in \calP \}} \left\{ |{\mathrm \mathit
    CPlans}(p_i)|\right\}.
\end{equation}

The plan $p_{max}$ is incrementally computed each time the local search finds a
new solution. In addition to being used to identify the reference plan
in $E$, $p_{max}$ is also used for defining the initial state (partial
plan) of the search process. Specifically, we initialize the search
using a (partial) plan obtained by randomly removing some actions from a
(randomly selected) plan in the set {\it CPlans}$(p_{max}) \cup
\{p_{max}\}$.

The process of generating diverse plans starting from a dynamically
chosen reference plan continues until at least $k$ plans that are all
$d$-distant from each other have been produced. The modified version of
\lpg\ to compute diverse plans is called \lpg-d.

\subsubsection{Experimental Analysis with LPG-d}
\label{sec:lpg-experiments}















\begin{figure*}[t]
\centering

\begin{tabular}{c}
\vspace{-0.2in} \includegraphics[angle=90,width=0.69\textwidth]{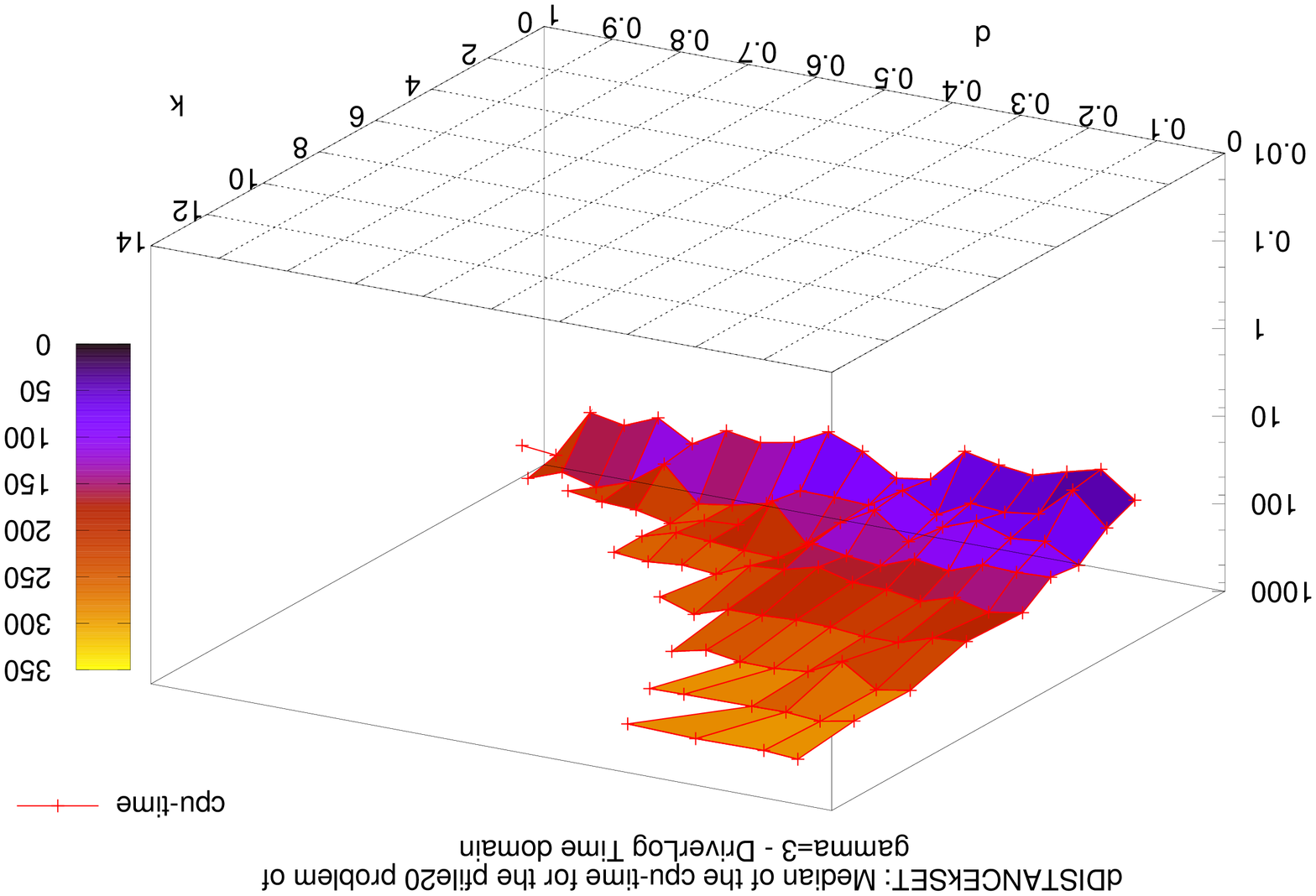} \\

\vspace{-0.15in} \includegraphics[angle=90,width=0.69\textwidth]{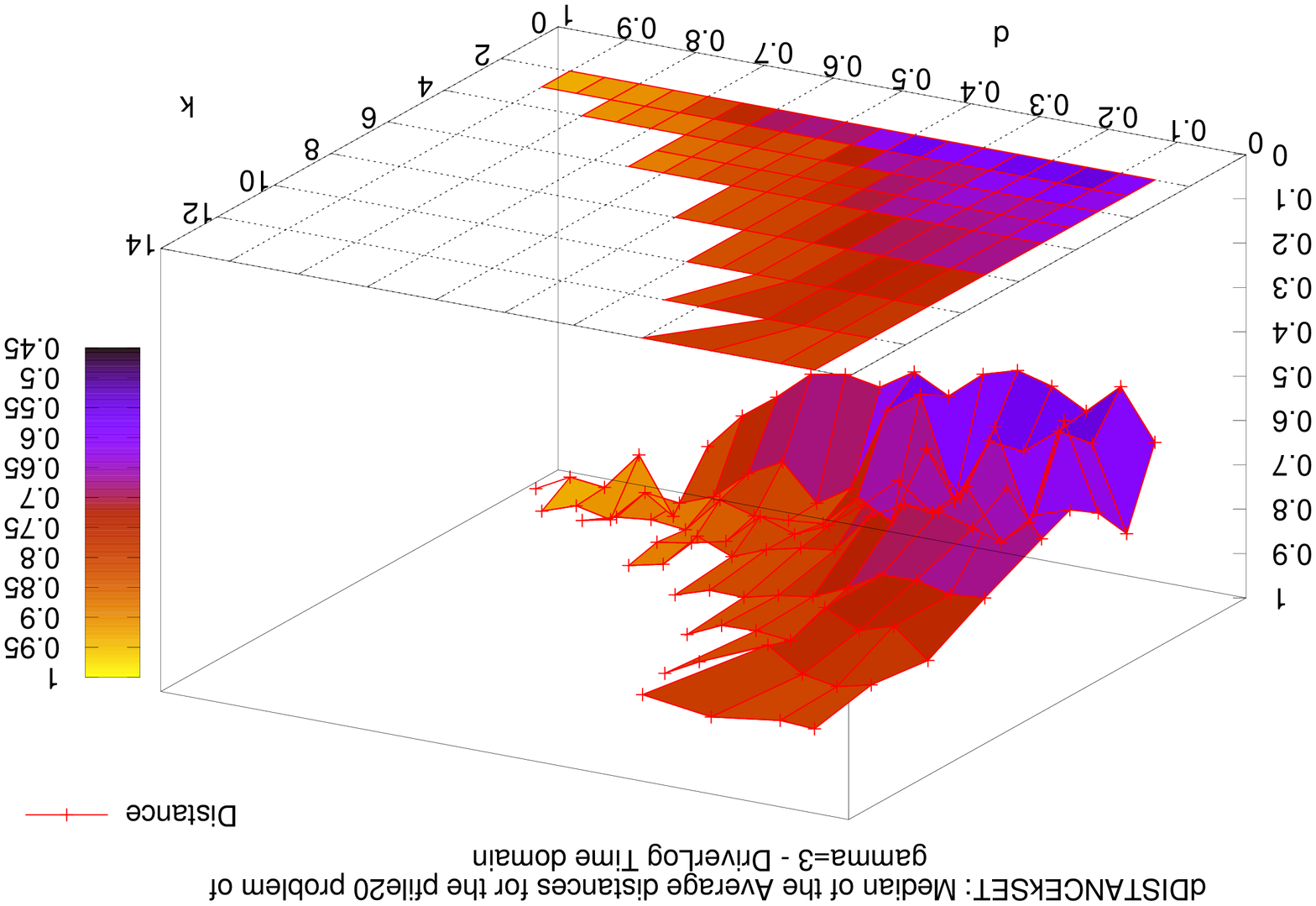}

\end{tabular}

\caption{Performance of \lpg-d (CPU-time and
  plan distance) for the problem pfile20 in the DriverLog-Time domain. }

\label{fig:driverlog-time}
\end{figure*}


\begin{figure*}[t]
\centering

\begin{tabular}{c}
\vspace{-0.2in}
\includegraphics[angle=90,width=0.69\textwidth]{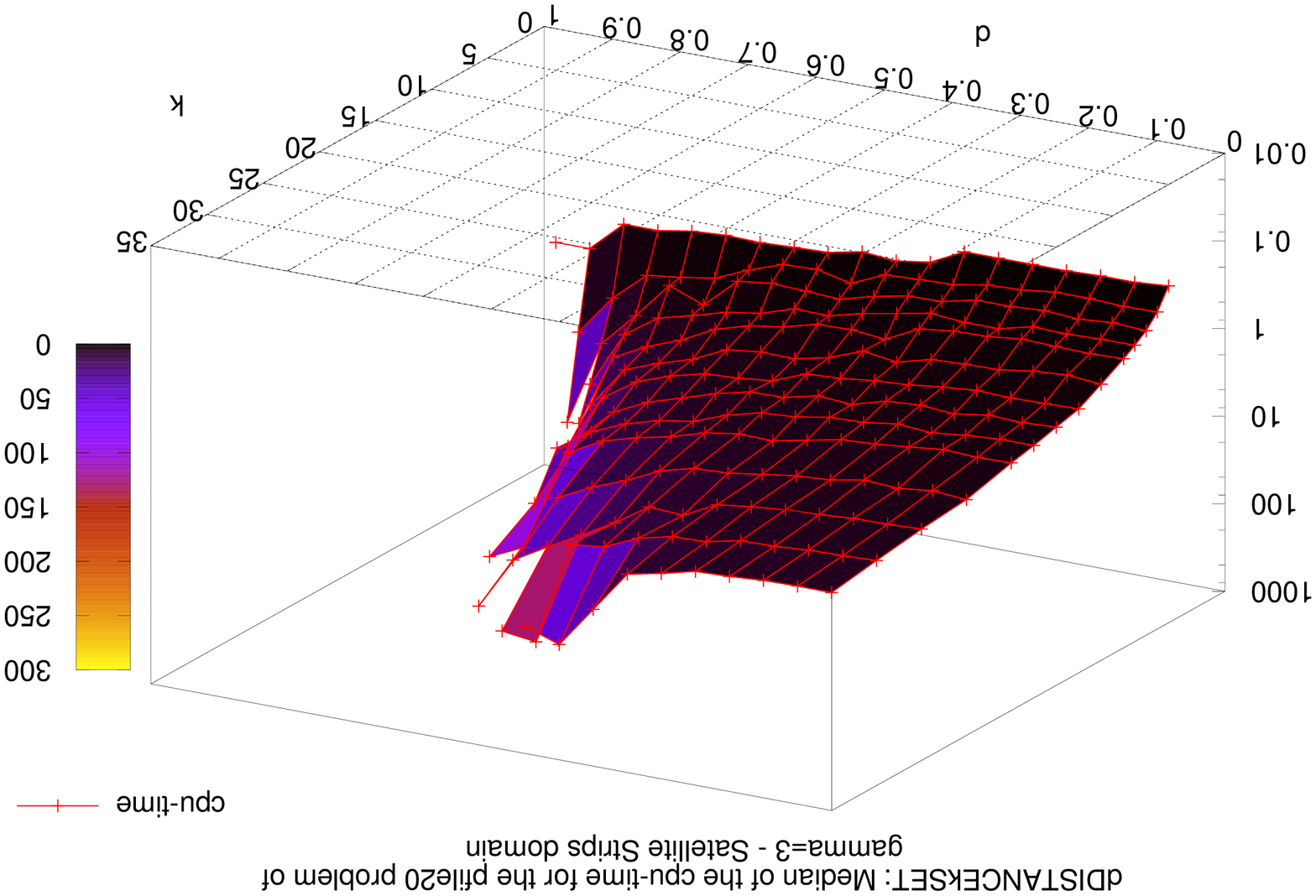} \\
\vspace{-0.15in}
\includegraphics[angle=90,width=0.69\textwidth]{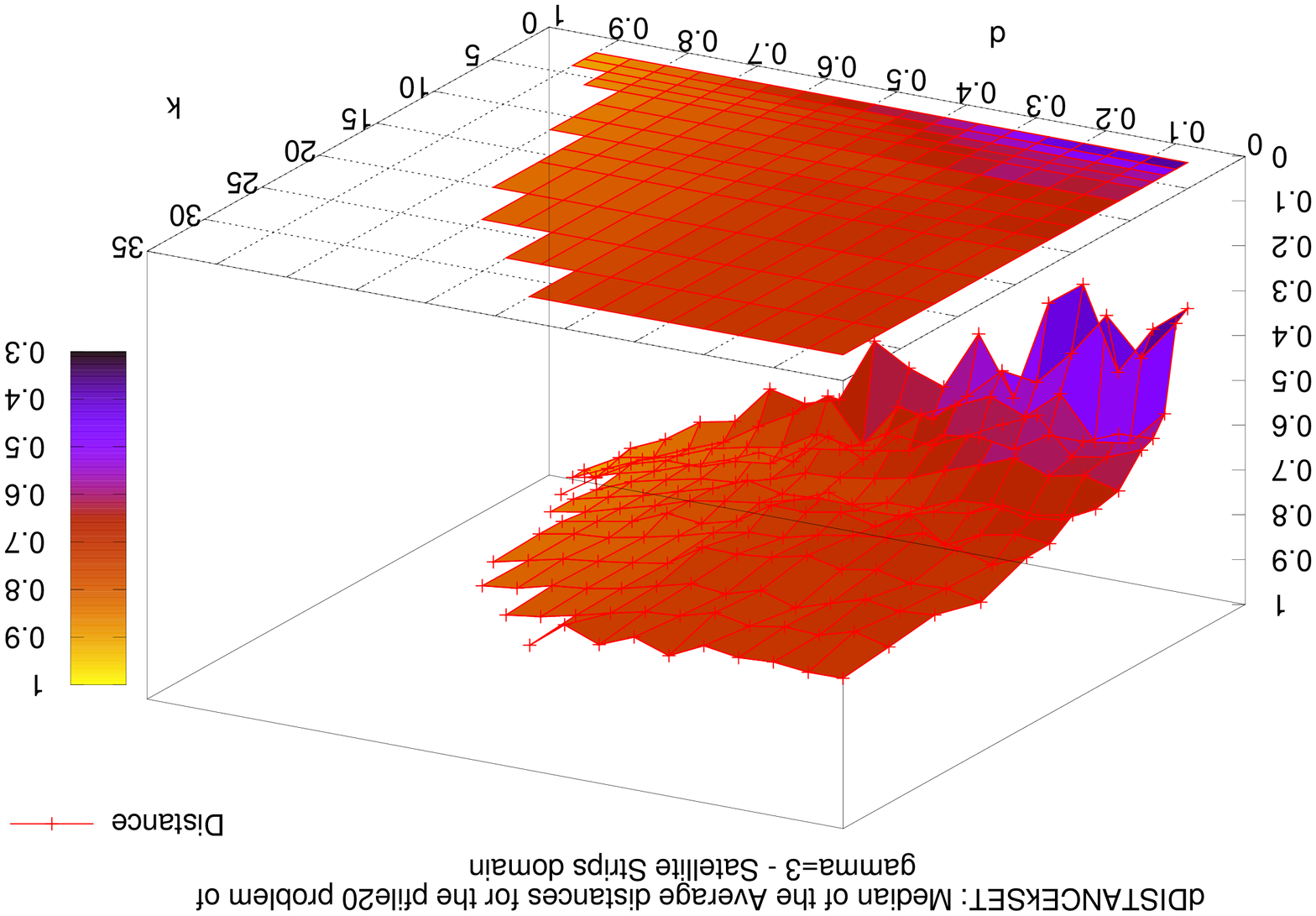} 

\end{tabular}

\caption{Performance of \lpg-d (CPU-time and
  plan distance) for the problem pfile20 in the Satellite-Strips domain. }

\label{fig:satellite-strips}
\end{figure*}


\begin{figure*}[t]
\centering

\begin{tabular}{c}
\vspace{-0.2in}
\includegraphics[angle=90,width=0.69\textwidth]{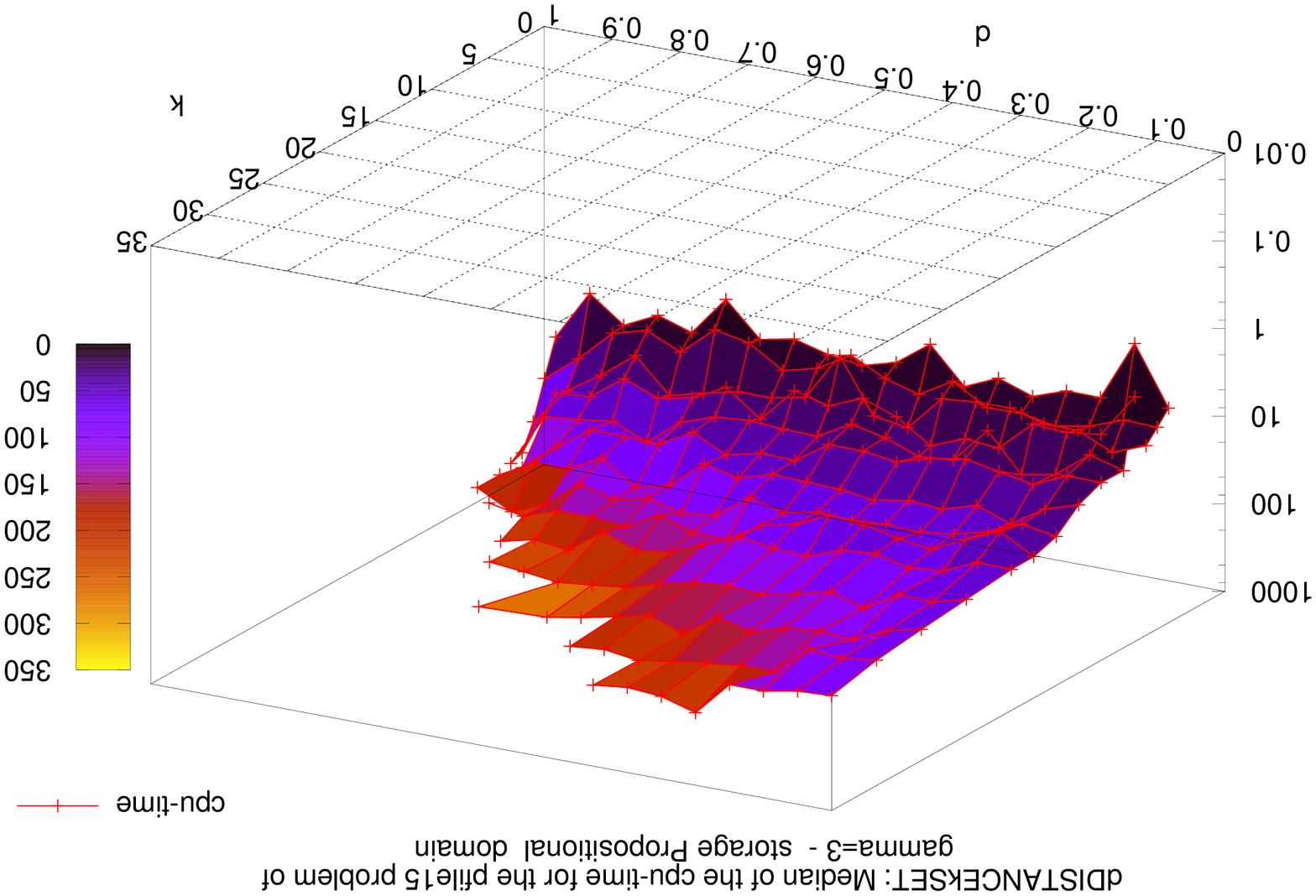} \\

\vspace{-0.15in}
\includegraphics[angle=90,width=0.69\textwidth]{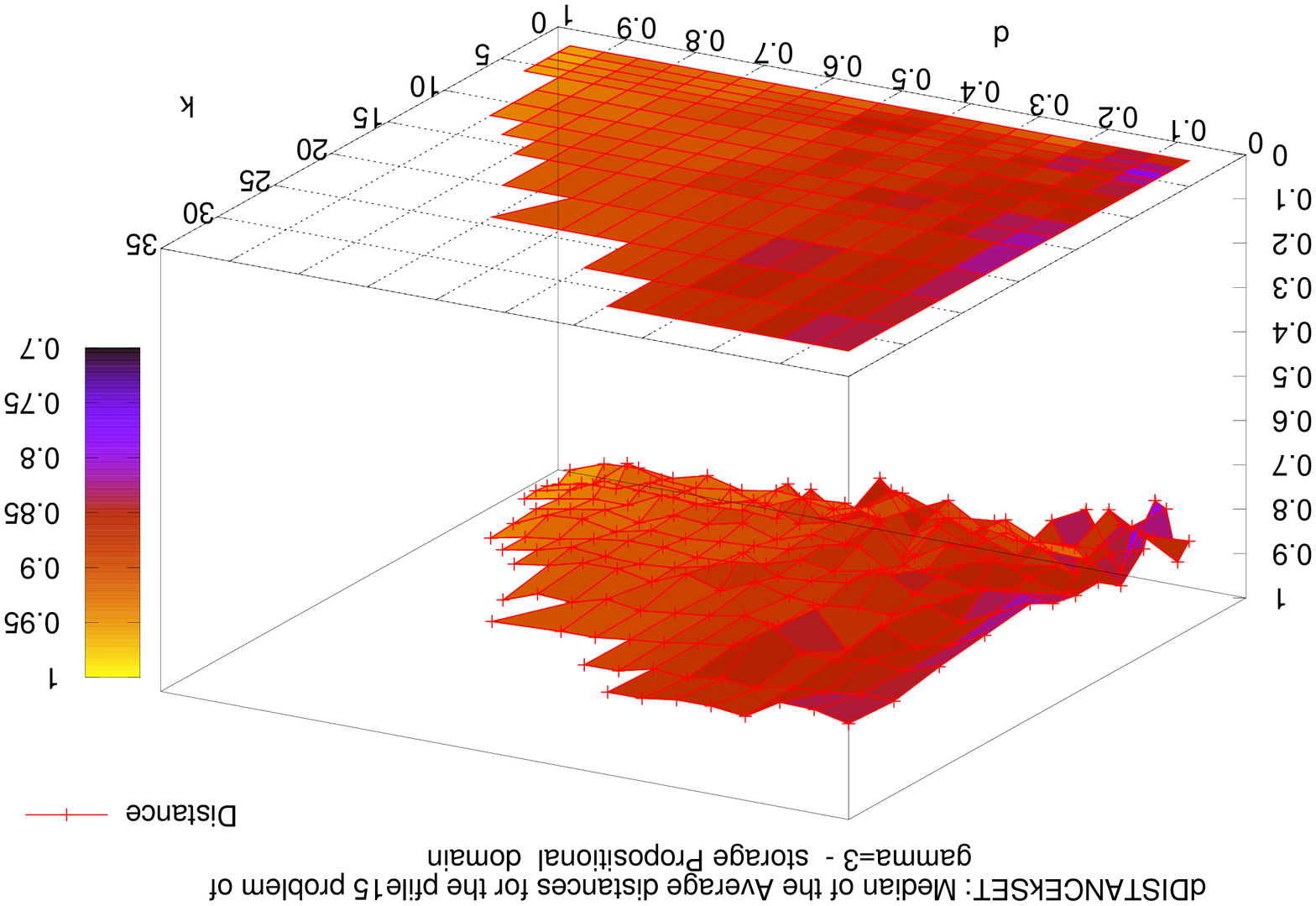}

\end{tabular}

\caption{Performance of \lpg-d (CPU-time and
  plan distance) for the problem pfile15 in the Storage-Propositional domain.}

\label{fig:storage-propositions}
\end{figure*}

\noindent
Recall that the distance function $\delta_a$, using set-difference, can be written
as the sum of two terms:

\begin{equation}
\delta_a(p_i, p_j)= \frac{|A(p_i) - A(p_j)| }{ |A(p_i) \cup A(p_j)|} + \frac{ |A(p_j) - A(p_i)|}{ |A(p_i) \cup A(p_j)|}
\end{equation}

The first term represents the contribution of the actions in $p_i$ to
the plan difference, while the second term indicates the contribution of
$p_j$ to $\delta_a$.
We experimentally observed that in some cases the differences between
two diverse plans computed using $\delta_a$ are mostly concentrated in
only one of the $\delta_a$ components. This asymmetry means that one of
the two plans can have many more actions than the other one, which could
imply that the quality of one of the two plans is much worse than the
quality of the other plan. In order to avoid this problem, we can
parametrize $\delta_a$ by imposing the two extra constraints
\begin{center} $\delta_a^A \ge d/\gamma$ and $\delta_a^B \ge d/\gamma$
\end{center}
\noindent where $\delta^A_a$ and $\delta^B_a$ are the first and second
terms of $\delta_a$, respectively, and $\gamma$ is an integer parameter
``balancing'' the diversity of $p_i$ and $p_j$.

In this section, we analyze the performance of the modified version of
\lpg, called \lpg-d, in three different benchmark domains from the 3rd
and 5th IPCs. The main goals of the experimental evaluation were (i)
showing that \lpg-d can efficiently solve a large set of
$(d,k)$-combinations, (ii) investigating the impact of the $\delta_a$
$\gamma$-constraints on performance, (iii) comparing \lpg-d and the
standard \lpg.

We tested \lpg-d using both the default and parametrized versions of
$\delta_a$, with $\gamma = 2$ and $\gamma = 3$.  We give detailed
results for $\gamma = 3$ and a more general evaluation for $\gamma = 2$
and the original $\delta_a$.  We consider $d$ that varies from $0.05$ to
$0.95$, using $0.05$ increment step, and with $k$ $=$ 2...5, 6, 8, 10,
12, 14, 16, 20, 24, 28, 32 (overall, a total of 266
$(d,k)$-combinations). Since \lpg-d is a stochastic planner, we use the
median of the CPU times (in seconds) and the median of the average plan
distances (over five runs). The average plan distance for a set of $k$
plans solving a specific $(d,k)$-combination ($\delta^{av}$) is the
average of the plans distances between all pairs of plans in the set.
The tests were performed on an AMD Athlon(tm) XP 2600+, 512 Mb RAM. The
CPU-time limit was 300 seconds.

Figure~\ref{fig:driverlog-time} gives the results for the
largest problem in IPC-3 DriverLog-Time (fully-automated
track).  \lpg-d solves $109$ $(d,k)$-combinations, including
combinations $d \leq 0.4$ and $k=10$, and $d=0.95$ and $k=2$.
The average CPU time (top plots) is $162.8$ seconds.  The average
$\delta^{av}$ (bottom plots) is $0.68$, with $\delta^{av}$ always
greater than $0.4$.  With the original $\delta_a$ function \lpg-d solves
$110$ $(d,k)$-combinations, the average CPU time is $160$ seconds, and
the average $\delta^{av}$ is $0.68$; while with $\gamma=2$ \lpg-d solves
$100$ combinations, the average CPU time is $169.5$ seconds, and the
average $\delta^{av}$ is $0.69$.

Figure~\ref{fig:satellite-strips} shows the results for the
largest problem in IPC-3 Satellite-Strips.  \lpg-d solves
$211$ $(k,d)$-combinations; $173$ of them require less than $10$
seconds. The average CPU time is $12.1$ seconds, and the average
$\delta^{av}$ is $0.69$. We observed similar results when using the
original $\delta_a$ function or the parametrized $\delta_a$
with~$\gamma=2$ (in the second case, \lpg-d solves 198 problems, while
the average CPU time and the average $\delta^{av}$ are nearly the same
as with $\gamma=3$).

Figure~\ref{fig:storage-propositions} shows the results for a
middle-size problem in IPC-5 Storage-Propositional. With $\gamma=2$ \lpg-d
solves $225$ $(k,d)$-combinations, $39$ of which require less than 10
seconds, while $128$ of them require less than 50 seconds. The average
CPU time is $64.1$ seconds and the average $\delta^{av}$ is $0.88$.
With the original $\delta_a$ \lpg-d solves $240$ $(k,d)$-combinations,
the average CPU time is $41.8$ seconds, and the average $\delta^{av}$ is
$0.87$; with $\gamma=3$ \lpg-d solves $206$ combinations, the average
CPU time is $69.4$ seconds and the average $\delta^{av}$ is $0.89$.



The local search in \lpg\ is randomized by a ``noise'' parameter that is
automatically set and updated during search \citep{lpg:JAIR03}. This
randomization is one of the techniques used for escaping local minima,
but it also can be useful for computing diverse plans: if we run the
search multiple times, each search is likely to consider different
portions of the search space, which can lead to different solutions.  It
is then interesting to compare \lpg-d and a method in which we simply
run the standard \lpg\ until $k$ $d$-diverse plans are generated.
An experimental comparison of the two approaches show that in many cases
\lpg-d performs better.  In particular, the new evaluation function $E$
is especially useful for planning problems that are easy to solve for
the standard \lpg, and that admit many solutions. In these cases, the
original $E$ function produces many valid plans with not enough
diversification.  This problem is significantly alleviated by the new
term in $E$.
An example of domain where we observed this behavior is {\small \tt
  logistics}.\footnote{E.g., for {\small \tt logistics\_a} (prob3
  of Table \ref{table:time_per_sol}) \lpg-d solved 128 instances, 41 of
  them in less than 1 CPU second and 97 of them in less than 10 CPU
  seconds; the average CPU time was $16.7$ seconds and the average
  $\delta^{av}$ was $0.38$. While using the standard \lpg, only 78
  instances were solved, 20 of them in less than 1 CPU seconds and 53 of
  them in less than 10 CPU seconds; the average CPU time was $23.6$
  seconds and the average $\delta^{av}$ was $0.27$.  }

\section{Generating Plan Sets with Partial Preference Knowledge}
\label{sec:planning-partial-preference}
\noindent
In this section, we consider the problem of generating plan sets when
the user's preferences are only partially expressed. In particular, we
focus on metric temporal planning where the preference model is assumed
to be represented by an incomplete value function specified by a convex combination of two features: \emph{plan
  makespan} and \emph{execution cost}, with the
exact trade-off value $w$ drawn from a given distribution. The quality value of plan sets is
measured by the ICP value, as formalized in Equation~\ref{eq:convex-combination}. Our objective is to find a set of
plans $\calP \subseteq \calS$ where $|\calP| \leq k$ and $ICP(\calP)$ is
the lowest.

Notice that we restrict the size of the solution set returned, not only
for the comprehension issue discussed earlier, but also for an important
property of the ICP measure: it is a monotonically non-increasing
function of the solution set (specifically, given two solution sets
${\cal P}_1$ and ${\cal P}_2$ such that the latter is a superset of
the former, it is easy to see that $ICP({\cal P}_2) \leq ICP({\cal
P}_1)$).


\subsection {Sampling Weight Values}
\label{subsec:sampling}

\noindent
Given that the distribution of trade-off value $w$ is known, the
straightforward way to find a set of representative solutions is to
first sample a set of $k$ values for $w$: $\{w_1,w_2,...,w_k\}$ based on
the distribution $h(w)$. For each value $w_i$, we can find an (optimal)
plan $p_i$ minimizing the value of the overall value function
$V(p,w_i) = w_i \times t_p + (1 - w_i) \times c_p$. The final set of
solutions $\calP = \{p_1,p_2,....p_k\}$ is then filtered to remove
duplicates and dominated solutions, thus selecting the plans making up the
lower-convex hull. The final set can then be returned to the user. While
intuitive and easy to implement, this sampling-based approach has
several potential flaws that can limit the quality of its resulting plan
set.

First, given that $k$ solution plans are searched sequentially and
independently of each other, even if the plan $p_i$ found for each
$w_i$ is optimal, the final solution set $\calP = \{p_1,p_2...p_k\}$
may not even be the optimal set of $k$ solutions with regard to the
ICP measure. More specifically, for a given set of solutions $\calP$,
some tradeoff value $w$, and two non-dominated plans $p$, $q$ such
that $V(p,w) < V(q,w)$, it is possible that $ICP(\calP \cup \{p\}) >
ICP(\calP \cup \{q\})$. In our running example (Figure~\ref{fig:icp_example}), let $\calP = \{p_2,
p_5\}$ and $w = 0.8$ then $V(p_1,w) = 0.8 \times 4 + 0.2 \times 25 =
8.2 < V(p_7,w) = 0.8 \times 12 + 0.2 \times 5 = 10.6$. Thus, the
planner will select $p_1$ to add to $\calP$ because it looks locally
better given the weight $w = 0.8$. However, $ICP(\{p_1,p_2,p_5\})
\approx 10.05 > ICP(\{p_2,p_5,p_7\}) \approx 7.71$ so indeed by taking
previous set into consideration then $p_7$ is a much better choice
than $p_1$.

Second, the values of the trade-off parameter $w$ are sampled based on a given
distribution, and independently of the particular planning problem
being solved. As there is no relation between the sampled $w$
values and the solution space of a given planning problem,
sampling approach may return very few distinct solutions even if
we sample a large number of weight values $w$. In our example, if all
$w$ samples have values $w \leq 0.67$ then the optimal solution
returned for any of them will always be $p_7$. However, we know that
$\calP^* = \{p_1, p_3, p_7\}$ is the optimal set according to the
$ICP$ measure. Indeed, if $w \leq 0.769$ then the sampling approach
can only find the set $\{p_7\}$ or $\{p_3,p_7\}$ and still not be able
to find the optimal set $\calP^*$.

\subsection {ICP Sequential Approach}
\label{subsec:icp-seq}

\noindent
Given the potential drawbacks of the sampling approach outlined above,
we also pursued an alternative approach that takes into account the ICP
measure more actively. Specifically, we incrementally build the solution
set $\calP$ by finding a solution $p$ such that $\calP \cup \{p\}$ has
the lowest ICP value. We can start with an empty solution set $\calP =
\emptyset$, then at each step try to find a new plan $p$ such that
$\calP \cup \{p\}$ has the lowest ICP value.

While this approach directly takes the ICP measure into consideration at
each step of finding a new plan and avoids the drawbacks of the
sampling-based approach, it also has its own share of potential flaws.
Given that the set is built incrementally, the earlier steps where the
first ``seed" solutions are found are very important.  The closer the
seed solutions are to the global lower convex hull, the better the
improvement in the ICP value. In our example (Figure~\ref{fig:icp_example}),
if the first plan found is $p_2$ then the subsequent plans found to best
extend $\{p_2\}$ can be $p_5$ and thus the final set does not come close
to the optimal set $\calP^* = \{p_1, p_3, p_7\}$.

\begin{algorithm}[t]
  {\bf Input:} A planning problem with a solution space $\cal{S}$;
  maximum number of plans required $k$; number of sampled trade-off
  values $k_0$ ($0 < k_0 < k$); time bound $t$;

{\bf Output}: A plan set $\calP$ ($|\calP| \leq k$)\;

\Begin{
$W \leftarrow$ sample $k_0$ values for $w$\;
$\calP \leftarrow$ find good quality plans in $\cal{S}$ for each $w \in W$\;
\While{ $|\calP| < k$ and $search\_{time} < t$ } {

  Search for $p$ s.t. $ICP(\calP \cup \{p\}) < ICP(\calP)$

  $\calP \leftarrow \calP \cup \{p\}$

} Return $\calP$ }
\caption{Incrementally find solution set $\calP$}
\label{alg:icp}
\end{algorithm}


\subsection {Hybrid Approach}
\label{subsec:hybrid}

\noindent
In this approach, we aim to combine the strengths of both the sampling
and ICP-sequential approaches. Specifically, we use sampling to find
several plans optimizing for different weights. The plans are then used
to seed the subsequent ICP-sequential runs. By seeding the hybrid
approach with good quality plan set scattered across the pareto optimal
set, we hope to gradually expand the initial set to a final set with a
much better overall ICP value. Algorithm~\ref{alg:icp} shows the
pseudo-code for the hybrid approach. We first independently sample the
set of $k_0$ values (with $k_0$ pre-determined) of $w$ given the
distribution on $w$ (step 4). We then run a heuristic planner multiple
times to find an optimal (or good quality) solution for each trade-off
value $w$ (step 5). We then collect the plans found and seed the
subsequent runs when we incrementally update the initial plan set with
plans that lower the overall ICP value (steps 6-8). The algorithm
terminates and returns the latest plan set (step 9) if $k$ plans are
found or the time bound exceeds.

\subsection{Making LPG Search Sensitive to ICP}
\label{subsec:lpg}

\noindent
Since the LPG planner used in the previous section cannot handle numeric
fluents, in particular the $totalcost$ representing plan cost that
we are interested in, we use a modified version of the Metric-LPG
planner \citep{metric-lpg} in implementing our
algorithms. Not only is Metric-LPG equipped with a very flexible
local-search framework that has been extended to handle various
objective functions, but also it can be made to search for
single or multiple solutions. Specifically, for the sampling-based
approach, we first sample the $w$ values based on a given distribution.
For each $w$ value, we set the metric function in the domain file to: $w
\times makespan + (1-w) \times totalcost$, and run the original LPG in
the quality mode to heuristically find the best solution within the time
limit for that metric function. The final solution set is filtered to remove any duplicate
solutions, and returned to the user.

For the ICP-sequential and hybrid approach, we can not use the original
LPG implementation as is and need to modify the neighborhood evaluation
function in LPG to take into account the ICP measure and the current
plan set $\calP$. For the rest of this section, we will explain this
procedure in detail.

\medskip
\noindent{\bf Background:}
Metric-LPG uses local search to find plans within the space of
\emph{numerical action graphs} (NA-graph). This leveled graph consists
of a sequence of interleaved proposition and action layers. The
proposition layers consist of a set of propositional and numerical
nodes, while each action layer consists of at most one action node, and
a number of no-op links. An NA-graph $G$ represents a valid plan if all
actions' preconditions are supported by some actions appearing in the
earlier level in $G$. The search neighborhood for each local-search step
is defined by a set of graph modifications to fix some remaining
inconsistencies (unsupported preconditions) $p$ at a particular level
$l$. This can be done by either inserting a new action $a$ supporting
$p$ or removing from the graph the action $a$ that $p$ is a precondition
of (which can introduce new inconsistencies).

Each local move creates a new NA-graph $G'$, which is evaluated as a
weighted combination of two factors: $SearchCost(G')$ and
$ExecCost(G')$.
%
Here, $SearchCost(G')$ is the amount of search effort to resolve
inconsistencies newly introduced by inserting or removing action $a$; it
is measured by the number of actions in a relaxed plan $R$ resolving all
such inconsistencies. The total cost $ExecCost(G')$, which is a default
function to measure plan quality, is measured by the total \emph{action
  execution costs} of all actions in $R$. The two weight adjustment
values $\alpha$ and $\beta$ are used to steer the search toward either
finding a solution quickly (higher $\alpha$ value) or better solution
quality (higher $\beta$ value). Metric-LPG then selects the local move leading
to the smallest $E(G')$ value.

\medskip
\noindent{\bf Adjusting the evaluation function $E(G')$ for finding set
  of plans with low ICP measure:} To guide Metric-LPG towards optimizing our
ICP-sensitive objective function instead of the original minimizing cost
objective function, we need to replace the default plan quality measure
$ExecCost(G')$ with a new measure $ICPEst(G')$.  Specifically, we adjust
the function for evaluating each new NA-graph generated by local moves
at each step to be a combination of $SearchCost(G')$ and $ICPEst(G')$.
%
Given the set of found plans $\calP = \{p_1, p_2, ..., p_n\}$,
$ICPEst(G')$ guides Metric-LPG's search toward a plan $p$ generated from $G'$
such that the resulting set $\calP \cup \{p\}$ has a minimum ICP value:
$p = \argmin\limits_{p} ICP(\calP \cup \{p\})$. Thus, $ICPEst(G')$
estimates the expected total ICP value if the best plan $p$ found by
expanding $G'$ is added to the current found plan set $\calP$. Like the
original Metric-LPG, $p$ is estimated by $p_R = G' \bigcup R$ where $R$
is the relaxed plan resolving inconsistencies in $G'$ caused by
inserting or removing $a$.  The $ICPEst(G')$ for a given NA-graph $G'$
is calculated as:
$ICPEst(G') = ICP(\calP \cup p_R)$
with the ICP measure as described in
Equation~\ref{eq:convex-combination}. Notice here that while $\calP$ is
the set of valid plans, $p_R$ is not. It is an invalid plan represented
by a NA-graph containing some unsupported preconditions. However,
Equation~\ref{eq:convex-combination} is still applicable as long as we
can measure the time and cost dimensions of $p_R$.  To measure the
makespan of $p_R$, we estimate the time points at which unsupported
facts in $G'$ would be supported in $p_R = G' \cup R$ and propagate them
over actions in $G'$ to its last level. We then take the earliest time
point at which all facts at the last level appear to measure the
makespan of $p_R$.  For the cost measure, we just sum the individual
costs of all actions in $p_R$.

At each step of Metric-LPG's local search framework, combining two
measures $ICPEst(G')$ and $SearchCost(G')$ gives us an evaluation function that fits right
into the original Metric-LPG framework and prefers a NA-graph $G'$ in
the neighborhood of $G$ that gives the best trade-off between the
estimated effort to repair and the estimated decrease in quality of
the next resulting plan set.

\subsection {Experimental Results}
\label{sec:result}

\begin{figure*}[t]
\centering
\epsfig{file=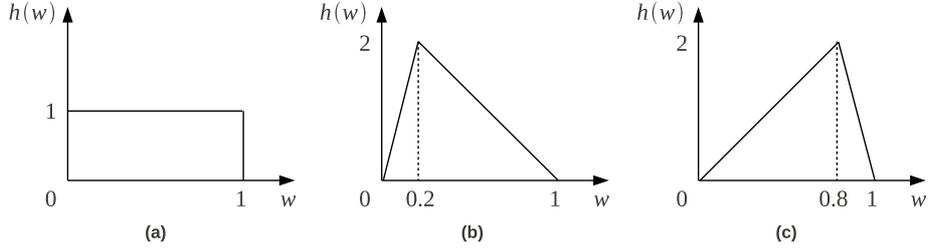,width=.9\linewidth}
\caption{The distributions: (a) uniform, (b) w02, (c) w08 (see text).}
\label{fig:distributions}
\end{figure*}

\noindent
We have implemented several approaches based on our algorithms discussed
in the previous sections: Sampling (Section~\ref{subsec:sampling}),
ICP-sequential (Section~\ref{subsec:icp-seq}) and Hybrid that combines
both (Section~\ref{subsec:hybrid}) with both the uniform and triangular
ditributions. We consider two types of distributions in which the most
probable weight for plan makespan are 0.2 and 0.8, which we will call
``w02'' and ``w08'' distributions respectively
(Figure~\ref{fig:distributions} shows these distributions). We test all
implementations against a set of 20 problems in each of several
benchmark temporal planning domains used in the previous International
Planning Competitions (IPC): \emph{ZenoTravel}, \emph{DriverLog}, and
\emph{Depots}. The only modification to the original benchmark set is
the added action costs.  The descriptions of these domains can be found
at the IPC website ( {\em ipc.icaps-conference.org}). The experiments
were conducted Intel Core2 Duo machine with 3.16GHz CPU and 4Gb RAM. For
all approaches, we search for a maximum of $k=10$ plans within the
10-minute time limit for each problem (i.e., $t=10$ minutes), and the resulting plan set is
used to compute the ICP value. In the Sampling approach, we generate ten
trade-off values $w$ between \emph{makespan} and \emph{plan cost} based
on the distribution, and for each one we search for a plan $p$ subject
to the value function $V(p,w)=w \times t_p + (1-w) \times c_p$. In the
Hybrid approach, on the other hand, the first Sampling approach is used
with $k_0 = 3$ generated trade-off values to find an initial plan
set, which is then improved by the ICP-Sequential runs. As Metric-LPG is a
stochastic local search planner, we run it three times for each problem
and average the results. In 77\% and 70\% of 60 problems in the three
tested domains for Hybrid and Sampling approaches respectively, the
standard deviation of ICP values of plan sets are at most 5\% of the
average values. This indicates that ICP values of plan set in different
runs are quite stable. As the Hybrid approach is an improved version of
ICP-sequential and gives better results in almost all tested problems,
we omit the ICP-Sequential in discussions below. We now analyze the
results in more detailed.

\begin{figure*}[t]
\centering
\epsfig{file=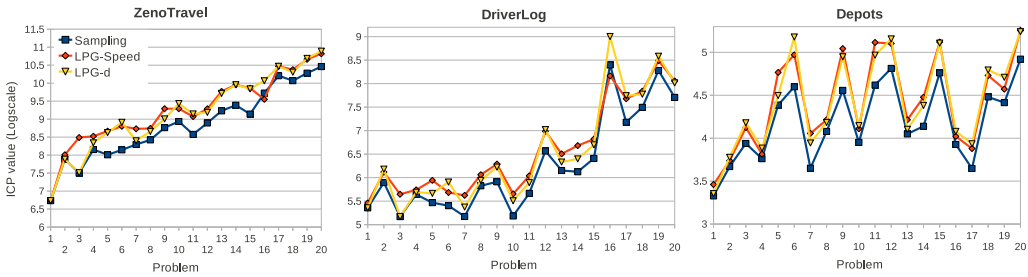,width=1.01\linewidth}
\caption{Results for the \emph{ZenoTravel}, \emph{DriverLog} and \emph{Depots} domains comparing the Sampling and
  baseline LPG approaches on the overall ICP value (log scale) with the
  uniform distribution.}
\label{fig:sampling_vs_naive}
\end{figure*}

\medskip
\noindent {\bf The utility of using the partial knowledge of user's
  preferences:} To evaluate the utility of taking partial preferences
into account, we first compare our results against the naive approaches
that generate a plan set without explicitly taking into account the
partial preference model. Specifically, we run the default LPG planner
with different random seeds to find multiple non-dominated plans. The
LPG planner was run with both \emph{speed} setting, which finds plans
quickly, and \emph{diverse} setting, which takes longer time to find
better set of diverse plans. Figure~\ref{fig:sampling_vs_naive} shows
the comparison between quality of plan sets returned by Sampling and
those naive approaches when the distribution of the trade-off value $w$
between \emph{makespan} and \emph{plan cost} is
assumed to be uniform.
Overall, among 20 tested problems for each of the ZenoTravel, DriverLog,
and Depots domains, the Sampling approach is better than LPG-speed in
19/20, 20/20 and 20/20 and is better than LPG-d in 18/20, 18/20, and
20/20 problems respectively. We observed similar results
comparing the Hybrid and those two approaches: in particular, the Hybrid
approach is better than LPG-speed in all 60 problems and better than
LPG-d in 19/20, 18/20, and 20/20 problems respectively. These results
support our intuition that taking into account the partial knowledge
about user's preferences (if it is available) increases the quality of
plan set.

\medskip
\noindent
{\bf Comparing the Sampling and Hybrid approaches:} We now compare the
effectiveness of the Sampling and Hybrid approaches in terms of the
quality of returned plan sets with the uniform, w02 and w08 distributions. 

\noindent
\textit {ICP value}: We first compare the two approaches in terms of the ICP values of plan
sets returned indicating their quality evaluated by the user.
Table~\ref{table:sampling_vs_hybrid_icp_zeno},~\ref{table:sampling_vs_hybrid_icp_driver},
and~\ref{table:sampling_vs_hybrid_icp_depots} show the results in three
domains ZenoTravel, DriverLog and Depots. In general,
Hybrid tends to be better than Sampling in this criterion for most of
the domains and distributions. In particular, in
ZenoTravel domain it returns higher quality plan sets in 15/20 problems
when the distribution is uniform, 10/20 and 13/20 problems when it is
w02 and w08 respectively (both approaches return plan sets with
equal ICP values for two problems with the w02 and one problem with the
w08 distribution). In the DriverLog domain, Hybrid returns better plan sets for 11/20
problems with the uniform distribution (and for other three problems the
plan sets have equal ICP values), but worse with the
triangular distributions: 8/20 (another 2 equals) and 9/20 (another one
equals) with w02 and w08. The improvement on the quality of plan sets that Hybrid
contributes is more significant in the Depots domain: it is better than
Sampling in 11/20 problems with the uniform distribution (and equal in 3
problems), in 12/20 problems with the w02 and w08 distributions (with
w02 both approaches return plan
sets with equal ICP values for 4 problems, and for 2 problems when it is w08).

\begin{table} {\footnotesize
\begin{center}
\begin{tabular}{| c | l | l || c | l | l || c | l | l | } 
\hline

{\bf Prob} & {\bf Sampling} & {\bf Hybrid} & {\bf Prob} & {\bf Sampling}
& {\bf Hybrid} & {\bf Prob} & {\bf Sampling} & {\bf Hybrid} \\
\hline \hline

1*     &     840.00     &     839.98     &     1     &     972.00     &     972.00     &     1     &     708.00     &     708.00    \\
2*     &     2,661.43     &     2,661.25     &     2     &     3,067.20     &     3,067.20     &     2*     &     2,255.792     &     2,255.788    \\
3*     &     1,807.84     &     1,805.95     &     3*     &     2,083.91     &     2,083.83     &     3*     &     1,535.54     &     1,535.32    \\
4*     &     3,481.31     &     3,477.49     &     4*     &     4,052.75     &     4,026.92     &     4*     &     2,960.84     &     2,947.66    \\
5*     &     3,007.97     &     2,743.85     &     5*     &     3,171.86     &     3,171.73     &     5*     &     2,782.16     &     2,326.94    \\
6*     &     3,447.37     &     2,755.25     &     6*     &     4,288.00     &     3,188.61     &     6*     &     2,802.00     &     2,524.18    \\
7*     &     4,006.38     &     3,793.44     &     7*     &     4,644.40     &     4,377.40     &     7*     &     3,546.95     &     3,235.63    \\
8*     &     4,549.90     &     4,344.70     &     8*     &     5,060.81     &     5,044.43     &     8*     &     3,802.60     &     3,733.90    \\
9*     &     6,397.32     &     5,875.13     &     9*     &     7,037.87     &     6,614.30     &     9*     &     5,469.24     &     5,040.88    \\
10*     &     7,592.72     &     6,826.60     &     10*     &     9,064.40     &     7,472.37     &     10*     &     6,142.68     &     5,997.45    \\
11*     &     5,307.04     &     5,050.07     &     11*     &     5,946.68     &     5,891.76     &     11*     &     4,578.09     &     4,408.36    \\
12*     &     7,288.54     &     6,807.28     &     12*     &     7,954.74     &     7,586.28     &     12     &     5,483.19     &     5,756.89    \\
13*     &     10,208.11     &     9,956.94     &     13*     &     11,847.13     &     11,414.88     &     13*     &     8,515.74     &     8,479.09    \\
14     &     11,939.22     &     13,730.87     &     14     &     14,474.00     &     15,739.19     &     14*     &     11,610.38     &     11,369.46    \\
15     &     9,334.68     &     13,541.28     &     15     &     16,125.70     &     16,147.28     &     15*     &     11,748.45     &     11,418.59    \\
16*     &     16,724.21     &     13,949.26     &     16     &     19,386.00     &     19,841.67     &     16     &     14,503.79     &     15,121.77    \\
17*     &     27,085.57     &     26,822.37     &     17     &     29,559.03     &     32,175.66     &     17     &     21,354.78     &     22,297.65    \\
18     &     23,610.71     &     25,089.40     &     18     &     28,520.17     &     29,020.15     &     18     &     20,107.03     &     21,727.75    \\
19     &     29,114.30     &     29,276.09     &     19     &     34,224.02     &     36,496.40     &     19     &     23,721.90     &     25,222.24    \\
20     &     34,939.27     &     37,166.29     &     20     &     39,443.66     &     42,790.97     &     20     &     28,178.45     &     28,961.51    \\

\hline

\multicolumn{1}{r}{} & \multicolumn{1}{c}{(a)} & \multicolumn{1}{r}{} & \multicolumn{1}{r}{} &
\multicolumn{1}{r}{(b)} & \multicolumn{1}{r}{} & \multicolumn{1}{r}{} & \multicolumn{1}{r}{(c)} &
\multicolumn{1}{r}{} \\

\end{tabular}

\caption{The ICP value of plan sets in ZenoTravel domain returned by the Sampling and
  Hybrid approaches with the distributions (a) uniform, (b) w02 and (c) w08. The
  problems where Hybrid returns plan sets with better quality than
  Sampling are marked with (*).}
\label{table:sampling_vs_hybrid_icp_zeno}
\end{center} }

\end{table}


\begin{table} {\footnotesize
\begin{center}
\begin{tabular}{| c | l | l || c | l | l || c | l | l | } 
\hline

{\bf Prob} & {\bf Sampling} & {\bf Hybrid} & {\bf Prob} & {\bf Sampling}
& {\bf Hybrid} & {\bf Prob} & {\bf Sampling} & {\bf Hybrid} \\
\hline \hline

1  	 &  	212.00  	 &  	212.00  	 &  	1  	 &  	235.99  	 &  	236.00  	 &  	1  	 &  	188.00  	 &  	188.00  	\\
2* 	& 	363.30 	& 	348.38 	& 	2* 	& 	450.07 	& 	398.46 	& 	2* 	& 	333.20 	& 	299.70 	\\
3 	& 	176.00 	& 	176.00 	& 	3 	& 	203.20 	& 	203.20 	& 	3 	& 	148.80 	& 	148.80 	\\
4* 	& 	282.00 	& 	278.45 	& 	4* 	& 	336.01 	& 	323.79 	& 	4* 	& 	238.20 	& 	233.20 	\\
5* 	& 	236.83 	& 	236.33 	& 	5 	& 	273.80 	& 	288.51 	& 	5* 	& 	200.80 	& 	199.52 	\\
6* 	& 	222.00 	& 	221.00 	& 	6 	& 	254.80 	& 	254.80 	& 	6* 	& 	187.47 	& 	187.20 	\\
7 	& 	176.50 	& 	176.50 	& 	7* 	& 	226.20 	& 	203.80 	& 	7 	& 	149.20 	& 	149.20 	\\
8* 	& 	338.96 	& 	319.43 	& 	8 	& 	387.53 	& 	397.75 	& 	8 	& 	300.54 	& 	323.87 	\\
9* 	& 	369.18 	& 	301.72 	& 	9* 	& 	420.64 	& 	339.05 	& 	9* 	& 	316.80 	& 	263.92 	\\
10* 	& 	178.38 	& 	170.55 	& 	10* 	& 	196.44 	& 	195.11 	& 	10* 	& 	158.18 	& 	146.12 	\\
11* 	& 	289.04 	& 	232.65 	& 	11* 	& 	334.13 	& 	253.09 	& 	11* 	& 	245.38 	& 	211.60 	\\
12 	& 	711.48 	& 	727.65 	& 	12* 	& 	824.17 	& 	809.93 	& 	12* 	& 	605.86 	& 	588.82 	\\
13* 	& 	469.50 	& 	460.99 	& 	13 	& 	519.92 	& 	521.05 	& 	13 	& 	388.80 	& 	397.67 	\\
14 	& 	457.04 	& 	512.11 	& 	14 	& 	524.56 	& 	565.94 	& 	14 	& 	409.02 	& 	410.53 	\\
15* 	& 	606.81 	& 	591.41 	& 	15* 	& 	699.49 	& 	643.72 	& 	15 	& 	552.79 	& 	574.95 	\\
16 	& 	4,432.21 	& 	4,490.17 	& 	16 	& 	4,902.34 	& 	6,328.07 	& 	16 	& 	3,580.32 	& 	4,297.47 	\\
17 	& 	1,310.83 	& 	1,427.70 	& 	17 	& 	1,632.86 	& 	1,659.46 	& 	17 	& 	1,062.03 	& 	1,146.68 	\\
18* 	& 	1,800.49 	& 	1,768.17 	& 	18 	& 	1,992.32 	& 	2,183.13 	& 	18 	& 	1,448.36 	& 	1,549.09 	\\
19 	& 	3,941.08 	& 	4,278.67 	& 	19 	& 	4,614.13 	& 	7,978.00 	& 	19* 	& 	3,865.54 	& 	2,712.08 	\\
20 	& 	2,225.66 	& 	2,397.61 	& 	20 	& 	2,664.00 	& 	2,792.90 	& 	20 	& 	1,892.28 	& 	1,934.11 	\\

\hline

\multicolumn{1}{r}{} & \multicolumn{1}{c}{(a)} & \multicolumn{1}{r}{} & \multicolumn{1}{r}{} &
\multicolumn{1}{r}{(b)} & \multicolumn{1}{r}{} & \multicolumn{1}{r}{} & \multicolumn{1}{r}{(c)} &
\multicolumn{1}{r}{} \\

\end{tabular}

\caption{The ICP value of plan sets in DriverLog domain returned by the Sampling and
  Hybrid approaches with the distributions (a) uniform, (b) w02 and (c) w08. The
  problems where Hybrid returns plan sets with better quality than
  Sampling are marked with (*).}
\label{table:sampling_vs_hybrid_icp_driver}
\end{center} }

\end{table}


\begin{table} {\footnotesize
\begin{center}
\begin{tabular}{| c | l | l || c | l | l || c | l | l | } 
\hline

{\bf Prob} & {\bf Sampling} & {\bf Hybrid} & {\bf Prob} & {\bf Sampling}
& {\bf Hybrid} & {\bf Prob} & {\bf Sampling} & {\bf Hybrid} \\
\hline \hline

1  	 &  	27.87  	 &  	27.87  	 &  	1  	 &  	28.56  	 &  	28.56  	 &  	1*  	 &  	28.50  	 &  	27.85  	\\
2 	& 	39.22 	& 	39.22 	& 	2 	& 	41.12 	& 	41.12 	& 	2 	& 	38.26 	& 	38.26 	\\
3* 	& 	51.36 	& 	50.43 	& 	3* 	& 	54.44 	& 	52.82 	& 	3* 	& 	49.49 	& 	48.58 	\\
4 	& 	43.00 	& 	43.00 	& 	4 	& 	46.00 	& 	46.00 	& 	4* 	& 	40.87 	& 	40.00 	\\
5 	& 	80.36 	& 	81.01 	& 	5 	& 	82.93 	& 	84.45 	& 	5 	& 	75.96 	& 	78.99 	\\
6 	& 	99.40 	& 	111.11 	& 	6 	& 	102.58 	& 	110.98 	& 	6 	& 	94.79 	& 	98.40 	\\
7* 	& 	38.50 	& 	38.49 	& 	7* 	& 	40.53 	& 	40.40 	& 	7* 	& 	37.04 	& 	36.60 	\\
8* 	& 	59.08 	& 	58.41 	& 	8* 	& 	62.15 	& 	62.08 	& 	8* 	& 	55.89 	& 	54.67 	\\
9 	& 	95.29 	& 	103.85 	& 	9 	& 	100.59 	& 	106.00 	& 	9 	& 	87.93 	& 	95.05 	\\
10* 	& 	52.04 	& 	50.00 	& 	10 	& 	52.40 	& 	52.40 	& 	10* 	& 	47.86 	& 	47.60 	\\
11 	& 	101.43 	& 	107.66 	& 	11* 	& 	110.18 	& 	108.07 	& 	11 	& 	97.56 	& 	99.06 	\\
12 	& 	123.09 	& 	129.34 	& 	12* 	& 	144.67 	& 	135.80 	& 	12 	& 	124.58 	& 	128.01 	\\
13* 	& 	57.37 	& 	57.22 	& 	13* 	& 	60.83 	& 	60.72 	& 	13 	& 	54.66 	& 	54.66 	\\
14* 	& 	62.75 	& 	62.33 	& 	14* 	& 	70.32 	& 	69.87 	& 	14* 	& 	65.20 	& 	62.02 	\\
15 	& 	116.82 	& 	117.86 	& 	15 	& 	113.15 	& 	124.28 	& 	15 	& 	101.09 	& 	124.43 	\\
16* 	& 	50.77 	& 	49.36 	& 	16* 	& 	54.98 	& 	54.12 	& 	16* 	& 	47.04 	& 	46.35 	\\
17* 	& 	38.38 	& 	37.77 	& 	17* 	& 	42.86 	& 	41.50 	& 	17* 	& 	37.56 	& 	36.92 	\\
18* 	& 	88.28 	& 	85.55 	& 	18* 	& 	94.53 	& 	90.02 	& 	18* 	& 	76.73 	& 	75.29 	\\
19* 	& 	82.60 	& 	82.08 	& 	19* 	& 	94.21 	& 	89.28 	& 	19* 	& 	74.73 	& 	72.45 	\\
20* 	& 	137.13 	& 	133.47 	& 	20* 	& 	150.80 	& 	135.93 	& 	20* 	& 	122.43 	& 	120.31 	\\

\hline

\multicolumn{1}{r}{} & \multicolumn{1}{c}{(a)} & \multicolumn{1}{r}{} & \multicolumn{1}{r}{} &
\multicolumn{1}{r}{(b)} & \multicolumn{1}{r}{} & \multicolumn{1}{r}{} & \multicolumn{1}{r}{(c)} &
\multicolumn{1}{r}{} \\

\end{tabular}

\caption{The ICP value of plan sets in Depots domain returned by the Sampling and
  Hybrid approaches with the distributions (a) uniform, (b) w02 and (c) w08. The
  problems where Hybrid returns plan sets with better quality than
  Sampling are marked with (*).}
\label{table:sampling_vs_hybrid_icp_depots}
\end{center} }

\end{table}

In many large problems of the ZenoTravel and DriverLog domains where
Sampling performs better than Hybrid, we notice that the first phase of
the Hybrid approach that searches for the first 3 initial plans normally
takes most of the allocated time, and therefore there is not much time
left for the second phase to improve the quality of plan set. We also
observe that among the three settings of the trade-off distributions,
the positive effect of the second phase in Hybrid approach (which is to
improve the quality of the initial plan sets) tends to be more stable
across different domains with uniform distribution, but less with the
triangular, in particular Sampling wins Hybrid in DriverLog domains when
the distribution is w02.  Perhaps this is because with the triangular
distributions, the chance that LPG planner (that is used in our
Sampling approach) returns the same plans even with different trade-off
values would increase, especially when the most probable value of
makespan happens to be in a (wide) range of weights in which one single
plan is optimal. This result agrees with the intuition that when the
knowledge about user's preferences is \emph{almost} complete (i.e. the
distribution of trade-off value is ``peak''), then Sampling approach
with smaller number of generated weight values may be good enough
(assuming that a good planner optimizing a complete value function is
available).

Since the quality of a plan set depends on how the two features makespan
and plan cost are optimized, and how the plans ``span'' the
space of time and cost, we also compare Sampling and Hybrid approaches in
terms of those two criteria. In particular, we compare plan sets returned
by the two approaches in terms of (i) their \emph{median} values of makespan and
cost, which represent how ``close'' the plan sets are to the origin of
the space of makespan and cost, and (ii) their \emph{standard deviation} of
makespan and cost values, which indicate how the sets span each feature axis.

\begin{table} 
\begin{center}{\footnotesize
\begin{tabular}{ l c | c | c | c | c |} 

\cline{3-6}

& & \multicolumn{2}{|c|}{\bf Median of makespan} &
\multicolumn{2}{|c|}{\bf Median of cost} \\
\hline

\multicolumn{1}{|c|}{\bf Domain} & \multicolumn{1}{|c|}{\bf Distribution} & $S>H$ & $H>S$ & $S>H$ & $H>S$ \\

\hline

\multicolumn{1}{|c|}{\multirow{3}{*}{ZenoTravel}} & uniform & 3 & {\bf
  17} & {\bf 16} & 4 \\

\multicolumn{1}{|c|}{} & w02 & 6 & {\bf 12} & {\bf 14} & 4 \\

\multicolumn{1}{|c|}{} & w08 & 6 & {\bf 13} & {\bf 13} & 6 \\

\hline

\multicolumn{1}{|c|}{\multirow{3}{*}{DriverLog}} & uniform & 6 & {\bf
  11} & 7 & {\bf 11} \\

\multicolumn{1}{|c|}{} & w02 & {\bf 10} & 8 & 8 & {\bf 10} \\

\multicolumn{1}{|c|}{} & w08 & {\bf 10} & 7 & 9 & 9 \\

\hline 

\multicolumn{1}{|c|}{\multirow{3}{*}{Depots}} & uniform & {\bf 9} & 8 &
{\bf 9} & 7 \\

\multicolumn{1}{|c|}{} & w02 & 7 & {\bf 9} & 5 & {\bf 9} \\

\multicolumn{1}{|c|}{} & w08 & {\bf 11} & 7 & 7 & {\bf 11} \\

\hline 

\end{tabular}}
\caption{The numbers of problems for each domain, distribution and
  feature where
  Sampling (Hybrid) returns plan sets with better (i.e. smaller) \emph{median}
  of feature value than that of Hybrid (Sampling), denoted in the table by $S>H$
  ($H>S$, respectively). We mark bold the numbers of problems that
  indicate the outperformance of the corresponding approach.}
\label{table:median}
\end{center}
\end{table}

Table~\ref{table:median} summarizes for each domain, distribution and
feature the number of problems in which each approach (either Sampling
or Hybrid) generates plan sets with better median of each feature value
(makespan and plan cost) than the other. There are 60 problems across 3
different distributions, so in total, 180 cases for each feature.
Sampling and Hybrid return plan sets with better makespan in 40 and 62
cases, and with better plan cost in 52 and 51 cases (respectively),
which indicates that Hybrid is slightly better than Sampling on
optimizing makespan but is possibly worse on optimizing plan cost. In
ZenoTravel domain, for all distributions Hybrid likely returns better
plan sets on the makespan than Sampling, and Sampling is better on the
plan cost feature. In DriverLog domain, Sampling is better on the
makespan feature with both non-uniform distributions, but worse than
Hybrid with the uniform. On the plan cost feature, Hybrid returns plan
sets with better median than Sampling on the uniform and w02
distribution, and both approaches perform equally well with the w08
ditribution. In Depots domain, Sampling is better than Hybrid on both
features with the uniform distribution, and only better than Hybrid on
the makespan with the distribution w08.

\begin{table} 
\begin{center}{\footnotesize
\begin{tabular}{ l c | c | c | c | c |} 

\cline{3-6}

& & \multicolumn{2}{|c|}{\bf SD of makespan} &
\multicolumn{2}{|c|}{\bf SD of cost} \\
\hline

\multicolumn{1}{|c|}{\bf Domain} & \multicolumn{1}{|c|}{\bf Distribution} & $S>H$ & $H>S$ & $S>H$ & $H>S$ \\

\hline

\multicolumn{1}{|c|}{\multirow{3}{*}{ZenoTravel}} & uniform & 8 & {\bf
  12} & 6 & {\bf 14} \\

\multicolumn{1}{|c|}{} & w02 & 4 & {\bf 14} & 7 & {\bf 11} \\

\multicolumn{1}{|c|}{} & w08 & 6 & {\bf 13} & 8 & {\bf 11} \\

\hline

\multicolumn{1}{|c|}{\multirow{3}{*}{DriverLog}} & uniform & 5 & {\bf
  11} & 6 & {\bf 10} \\

\multicolumn{1}{|c|}{} & w02 & 7 & {\bf 10} & 7 & {\bf 9} \\

\multicolumn{1}{|c|}{} & w08 & 8 & {\bf 9} & {\bf 10} & 7 \\

\hline 

\multicolumn{1}{|c|}{\multirow{3}{*}{Depots}} & uniform & {\bf 10} & 7 &
7 & {\bf 9} \\

\multicolumn{1}{|c|}{} & w02 & 7 & {\bf 9} & 5 & {\bf 10} \\

\multicolumn{1}{|c|}{} & w08 & 5 & {\bf 13} & 7 & {\bf 11} \\

\hline 

\end{tabular}}
\caption{The numbers of problems for each domain, distribution and
  feature where
  Sampling (Hybrid) returns plan sets with better (i.e. larger)
  \emph{standard deviation} of feature value than that of Hybrid (Sampling), denoted in the table by $S>H$
  ($H>S$, respectively). We mark bold the numbers of problems that
  indicate the outperformance of the corresponding approach.}
\label{table:standard-deviation}
\end{center}
\end{table}


In terms of spanning plan sets, Hybrid performs much better than
Sampling on both features across three domains, as shown in
Table~\ref{table:standard-deviation}. In particular, over 360 cases for
both makespan and plan cost features, there are only 10 cases where
Sampling produces plan sets with better standard deviation than Hybrid
on each feature. Hybrid, on the other hand, generates plan sets with
better standard deviation on makespan in 91 cases, and in 85 cases on
the plan cost.

These experimental results support our arguments in
Section~\ref{subsec:sampling} about the limits of sampling idea. Since
one single plan could be optimal for a wide range of weight values, the
search in Sampling approach with different trade-off values may focus on
looking for plans only at the same region of the feature space
(specified by the particular value of the weight), which can reduce the
chance of having plans with better value on some particular feature. On
the opposite side, the Hybrid approach tends to be better in spanning
plan sets to a larger region of the space, as the set of plans that have
been found is taken into account during the search.

\begin{figure*}[t]
\centering
\epsfig{file=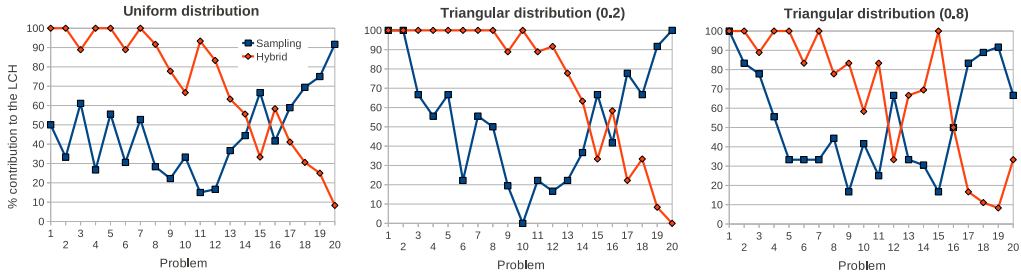,width=1.03\linewidth}
\caption{The contribution into the
  common lower convex hull of plan sets in ZenoTravel domain with different distributions.}
\label{fig:sampling_vs_hybrid_lch_zeno}
\end{figure*}

\begin{figure*}[t]
\centering
\epsfig{file=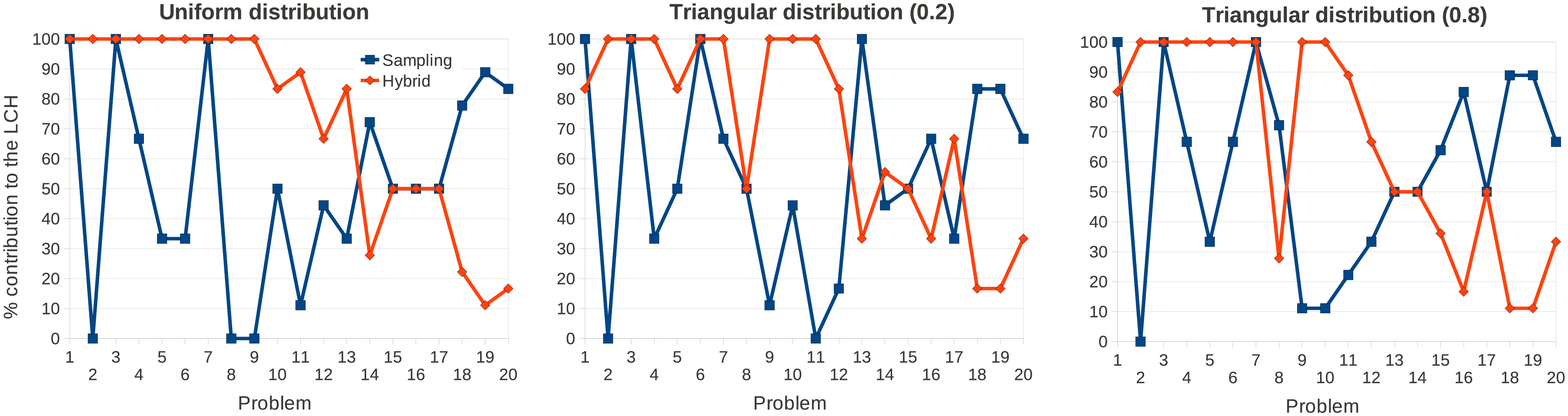,width=1.03\linewidth}
\caption{The contribution into the
  common lower convex hull of plan sets in DriverLog domain with different distributions.}
\label{fig:sampling_vs_hybrid_lch_driver}
\end{figure*}

\begin{figure*}[t]
\centering
\epsfig{file=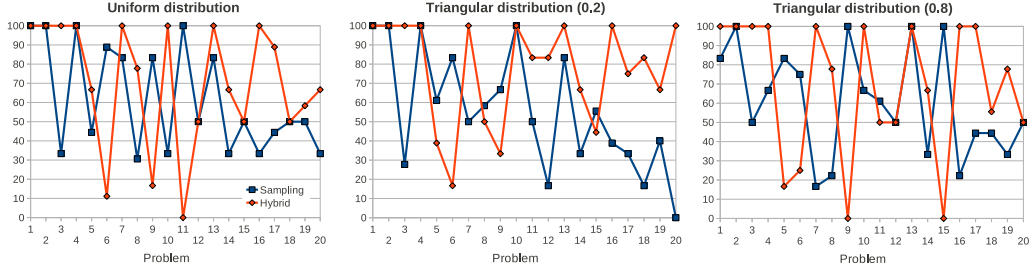,width=1.01\linewidth}
\caption{The contribution into the
  common lower convex hull of plan sets in Depots domain with different distributions.}
\label{fig:sampling_vs_hybrid_lch_depots}
\end{figure*}

\medskip
\noindent
\textit{Contribution to the lower convex hull}: The
comparison above between Sampling and Hybrid considers the two features
separately. We now examine the relation between plan sets returned by
those approaches on the joint space of both features, in particular taking
into account the the dominance relation between plans in the two
sets. In other words, we compare the relative total number of plans in the lower
convex-hull (LCH) found by each approach. Given that this is the set that
should be returned to the user (to select one from), the higher number
tends to give her a better expected utility value. To measure the
relative performance of both approaches with respect to this criterion, we first create a set $S$
combining the plans returned by them. We then
compute the set $S_{lch} \subseteq S$ of plans in the lower convex hull
among all plans in $S$. Finally, we measure the percentages of plans in $S_{lch}$
that are actually returned by each of our tested approaches. 
Figures~\ref{fig:sampling_vs_hybrid_lch_zeno},~\ref{fig:sampling_vs_hybrid_lch_driver}
and~\ref{fig:sampling_vs_hybrid_lch_depots} show the contribution to
the LCH of plan sets returned by Sampling and
Hybrid in ZenoTravel, DriverLog and Depots domains.


In general, we observe that the plan set returned by Hybrid contributes
more into the LCH than that of Sampling for most of the problems (except
for some large problems) with
most of the distributions and domains. Specifically, in ZenoTravel
domain, Hybrid contributes more plans to the LCH than Sampling in 15/20,
13/20 (and another 2 equals), 13/20 (another 2 equals) problems for the
uniform, w02 and w08 distributions respectively. In DriverLog domain, it
is better than Sampling in 10/20 (another 6 equals), 10/20 (another 4
equals), 8/20 (another 5 equals) problems; and Hybrid is better in 11/20
(another 6 equals), 11/20 (another 4 equals) and 11/20 (another 4
equals) for the uniform, w02 and w08 distributions in Depots domain.
Again, similar to the ICP value, the Hybrid approach is less
effective on problems with large size (except with the w08 distribution
in Depots domain) in which the searching time is mostly used for finding
initial plan sets. We also note that a plan set with higher contribution
to the LCH is \emph{not} guaranteed to have better quality, except for
the extreme case where one plan set contributes 100\% and completely
dominates the other which contributes 0\% to the LCH. For example,
consider the problem 14 in ZenoTravel domain: even though the plan sets
returned by Hybrid contribute more than those of Sampling in all three
distributions, it is only the w08 where it has a better ICP value. The
reason for this is that the ICP value depends also on the range of the
trade-off value (and its density) for which a plan in the LCH is
optimal, whereas the LCH is constructed by simply comparing plans in
terms of their makespan and cost separately (i.e. using the dominance
relation), ignoring their relative importance.

\medskip
\noindent {\bf The sensitivity of plan sets to the distributions:} All
analysis having been done so far is to compare the effectiveness of
approaches with respect to a particular distribution of the trade-off
value. In this part, we examine how sensitive the plan sets are with
respect to different distributions. 

\noindent
\textit{Optimizing high-priority feature}: We first consider how plan
sets are optimized on each feature (makespan and plan cost) by each
approach with respect to two non-uniform distributions w02 and w08.
Those are the distributions representing scenarios where the users have
different priority on the features, and plan sets should be biased
to optimizing the feature that has higher priority (i.e. larger value of
weight). In particular, plans generated using the w08 distribution should
have better (i.e. \emph{smaller}) makespan values than those found with
the w02 distribution (since in the makespan has higher priority in w08
than it is in w02); on the other hand, plan set returned with w02 should
have better values of plan cost than those with w08.

\begin{table}
\begin{center}{\footnotesize
\begin{tabular}{ l c | c | c || c | c |} 

\cline{3-6}

& & \multicolumn{2}{|c|}{\bf Median of makespan} &
\multicolumn{2}{|c|}{\bf Median of cost} \\
\hline

\multicolumn{1}{|c|}{\bf Approach} & \multicolumn{1}{|c|}{\bf Domain} & $w02>w08$ & $w08>w02$ & $w02>w08$ & $w08>w02$ \\

\hline

\multicolumn{1}{|c|}{\multirow{3}{*}{Sampling}} & ZenoTravel & {\bf 5} & {\bf
  13} & {\bf 11} & {\bf 8} \\

\multicolumn{1}{|c|}{} & DriverLog & {\bf 6} & {\bf 10} & {\bf 13} &
{\bf 5} \\

\multicolumn{1}{|c|}{} & Depots & {\bf 6} & {\bf 12} & {\bf 10} & {\bf 7} \\

\hline

\multicolumn{1}{|c|}{\multirow{3}{*}{Hybrid}} & ZenoTravel & {\bf 5} & {\bf
  10} & {\bf 10} & {\bf 4} \\

\multicolumn{1}{|c|}{} & DriverLog & {\bf 4} & {\bf 10} & 6 & 9 \\

\multicolumn{1}{|c|}{} & Depots & {\bf 8} & {\bf 10} & 4 & 11 \\

\hline 

\end{tabular}}
\caption{The number of problems for each approach, domain and
  feature where plan sets returned with the w02 (w08) distribution with better (i.e. smaller)
  \emph{median} of feature value than that with w08 (w02), denoted in the table by $w02>w08$
  ($w08>w02$, respectively). For each approach, we mark bold the numbers
for domains in which there are more problems whose plan sets returned
with w08 (w02) have better makespan (plan cost) median than those
with w02 (w08, respectively).}
\label{table:median-2}
\end{center}
\end{table}

Table~\ref{table:median-2} summarizes for each domain, approach and
feature, the number of problems in which plan sets returned with one
distribution (either w02 or w08) have better \emph{median} value than with the
other. 
We
observe that for both features, the Sampling approach is very likely
to ``push'' plan sets
to regions of the space of makespan and cost with better value of
more interested feature. On the other hand, the Hybrid approach tends to
be more sensitive to the distributions on both the features in
ZenoTravel domain, and is more sensitive only on the
makespan feature in DriverLog and Depots domain. Those results generally
show that our approaches can bias the search towards optimizing features
that are more desired by the user.

\medskip
\noindent
\textit{Spanning plan sets on individual features}: Next, we examine how
plan sets span each feature, depending on the
degree of incompleteness of the distributions. Specifically, we compare the
\emph{standard deviation} of plan sets returned using the uniform
distribution with those generated using the distributions w02 and
w08. Intuitively, we expect that plan sets returned with the uniform
distribution would have higher standard deviation than those with the
distributions w02 and w08.

\begin{table} 
\begin{center}{\footnotesize
\begin{tabular}{ l c | c | c || c | c |} 

\cline{3-6}

& & \multicolumn{2}{|c|}{\bf SD of makespan} &
\multicolumn{2}{|c|}{\bf SD of cost} \\
\hline

\multicolumn{1}{|c|}{\bf Approach} & \multicolumn{1}{|c|}{\bf Domain} & $U>w02$ & $w02>U$ & $U>w02$ & $w02>U$ \\

\hline

\multicolumn{1}{|c|}{\multirow{3}{*}{Sampling}} & ZenoTravel & 9 & 10 & {\bf 10} & {\bf 7} \\

\multicolumn{1}{|c|}{} & DriverLog & 6 & 8 & 7 & 8 \\

\multicolumn{1}{|c|}{} & Depots & {\bf 9} & {\bf 6} & {\bf 8} & {\bf 7} \\

\hline

\multicolumn{1}{|c|}{\multirow{3}{*}{Hybrid}} & ZenoTravel & 9 & 10 & {\bf 12} & {\bf 7} \\

\multicolumn{1}{|c|}{} & DriverLog & 6 & 9 & {\bf 8} & {\bf 7} \\

\multicolumn{1}{|c|}{} & Depots & {\bf 8} & {\bf 6} & {\bf 9} & {\bf 4} \\

\hline 

\end{tabular}}
\caption{The numbers of problems for each approach, domain and
  feature where plan sets returned with the uniform (w02) distribution have better (i.e. higher)
  \emph{standard deviation} of the feature value than that with w02 (uniform), denoted in the table by $U>w02$
  ($w02>U$, respectively). For each approach and feature, we mark bold the numbers
for domains in which there are more problems whose plan sets returned
with the uniform distribution have better standard deviation value of
the feature than those
with the w02 distribution.}
\label{table:standard-deviation-w02}
\end{center}
\end{table}


\begin{table} 
\begin{center}{\footnotesize
\begin{tabular}{ l c | c | c || c | c |} 

\cline{3-6}

& & \multicolumn{2}{|c|}{\bf SD of makespan} &
\multicolumn{2}{|c|}{\bf SD of cost} \\
\hline

\multicolumn{1}{|c|}{\bf Approach} & \multicolumn{1}{|c|}{\bf Domain} & $U>w08$ & $w08>U$ & $U>w08$ & $w08>U$ \\

\hline

\multicolumn{1}{|c|}{\multirow{3}{*}{Sampling}} & ZenoTravel & {\bf 11}
& {\bf 8} & {\bf 15} & {\bf 4} \\

\multicolumn{1}{|c|}{} & DriverLog & 5 & 10 & 5 & 9 \\

\multicolumn{1}{|c|}{} & Depots & {\bf 12} & {\bf 7} & {\bf 12} & {\bf 6} \\

\hline

\multicolumn{1}{|c|}{\multirow{3}{*}{Hybrid}} & ZenoTravel & {\bf 10} &
{\bf 9} & {\bf 15} & {\bf 4} \\

\multicolumn{1}{|c|}{} & DriverLog & 7 & 7 & {\bf 8} & {\bf 6} \\

\multicolumn{1}{|c|}{} & Depots & 5 & 8 & {\bf 11} & {\bf 4} \\

\hline 

\end{tabular}}
\caption{The numbers of problems for each approach, domain and feature
  where plan sets returned with the uniform (w08) distribution with
  better (i.e. higher) \emph{standard deviation} of feature value than
  that with w08 (uniform), denoted in the table by $U>w08$
  ($w08>U$, respectively). For each approach and feature, we mark bold the numbers
  for domains in which there are more problems whose plan sets returned
  with the uniform distribution have better standard deviation value of
  the feature than those with the w08 distribution.}
\label{table:standard-deviation-w08}
\end{center}
\end{table}

Table~\ref{table:standard-deviation-w02} shows for each approach, domain
and feature, the number of problems generated with the uniform
distribution that have better standard deviation on the feature than
those found with the distribution w02. We observe that with the makespan
feature, both approaches return plan sets that are more ``spanned'' on
makespan in the Depots domain, but not with ZenoTravel and DriverLog.
With the plan cost feature, Hybrid shows its positive impact on all
three domains, whereas Sampling shows it with the ZenoTravel and Depots
domain. Similarly, table~\ref{table:standard-deviation-w08} shows the
results comparing the uniform and w08 distributions. This time, Sampling
returns plan sets with better standard deviation on both features in the
ZenoTravel and Depots domains, but not in DriverLog. Hybrid also shows
this in ZenoTravel domain, but for the remaining two domains, it tends
to return plan sets with expected standard deviation on the plan cost
feature only. From all of these results, we observe that with the
uniform distribution, both approaches likely generate plan sets that 
span better than with non-uniform distributions, especially on the
plan cost feature.

In summary, the experimental results in this section support the
following hypotheses:

\begin{itemize}

\item Instead of ignoring user's preference models which are partially
  specified, one should take them into account during plan generation, as
  plan sets returned would have better quality.

\item In generating plan sets sequentially to cope with partial
  preference models, Sampling approach that searches for plans
  separately and independently of the solution space tends to return worse quality
  plan sets than Hybrid approach.

\item The resulting plan sets returned by Hybrid approach tend to be more
  sensitive to the user's preference models than those found by Sampling approach.

\end{itemize}

\section{Related Work}
\label{sec:related-work}

\noindent
Currently there are very few research efforts in the planning literature
that explicitly consider incompletely specified user preferences during
planning. The usual approach for handling multiple objectives is to
assume that a specific way of combining the objectives is available
\citep{refanidis2003multiobjective,sapa}, and search for one optimal
plan with respect to this function. Brafman \&
Chernyavsky (\citeyear{brafman05a}) discuss a CSP-based approach to find
a plan for the most prefered goal state given the qualitative
preferences on goals. There is no action cost and makespan measurements
such as in our problem setting. Other relevant work includes
~\cite{bryce07}, in which the authors devise a variant of LAO* algorithm
to search for a conditional plan with multiple execution options for
each observation branch that are non-dominated with respect to
objectives like probability and cost to reach the goal.



In the context of decision-theoretic planning, some work has been
focused on scenarios where the value function is not completely defined,
in particular due to the incompleteness in specifying the reward
function. In those cases, one approach is to search for the most robust
policy with different robustness criteria (e.g.,
\citealp*{delage2007percentile,regan2010robust,nilim2005robust}). The
idea of searching for sets of policies has also been considered recently
in reinforcement learning. Specifically, in \citep{natarajan2005dynamic}
the reward function is incomplete with weight values changing over time,
and a set of policies is searched and stored so that whenever the
weights change a new best policy can be found by improving those in the
set. On the other hand,~\cite{barrett2008learning} provide Bellman
equations for the Q-values using all vectors on convex hull to search
for the whole pareto set.

Our work on planning with partial user's preferences is also related to
work on preference elicitation and decision making under uncertainty of
preferences. For instance, Chajewska et al
(\citeyear{chajewska2000making}) consider a decision making scenario
where the utility function is assumed to be drawn from a known
distribution, and either a \emph{single} best strategy or an elicitation
question will be suggested based on the expected utility of the strategy
and the value of information of the question. Boutilier et al.
(\citeyear{boutilier2010}) considers preference elicitation problem in
which the user's preference model is incomplete on both the set of
features and the utility function. However, these scenarios are
different from ours in two important issues: we focus on efficient
approach to synthesizing plans with respect to the partial preferences,
whereas the ``outcomes'' or ``configurations'' in their cases are
considered given upfront (or could be obtained with low cost), and we
aim to search for a set of plans based on a quality measure of plan sets
(instead of a quality measure over individual outcome or configuration).

Our approach to generating diverse plan sets to cope with planning
scenarios without knowledge of user's preferences is in the same spirit
as \citep{tate1998} and Myers \citep{metatheory,metatheory-diverse},
though for different purposes. Myers, in particular, presents an
approach to generate diverse plans in the context of her HTN planner by
requiring the meta-theory of the domain to be available and using bias
on the meta-theoretic elements to control
search~\citep{metatheory-diverse}. The metatheory of the domain is
defined in terms of pre-defined attributes and their possible values
covering roles, features and measures. Our work differs from hers in two
respects. First, we focus on domain-independent distance measures.
Second we consider the computation of diverse plans in the context of
state of the art domain independent planners.

The problem of finding multiple but similar plans has been considered in the
context of replanning. A recent effort in this direction is 
 \citep{fox2006plan}. Our work focuses on the problem of finding diverse plans
by a variety of distance measures when the user's preferences exist but
are completely unknown.

Outside the planning literature, our closest connection is to the
work by Hebrard et al. \citeyear{csp-diversity}, who solve the 
problem of finding similar/dissimilar solutions for CSPs without additional
domain knowledge. It is instructive to note that unlike CSP, where the number of
potential solutions is finite (albeit exponential), the number of
distinct plans for a given problem can be infinite (since we can have
infinitely many non-minimal versions of the same plan). Thus,
effective approaches for generating diverse plans are even more critical.
The challenges in  finding interrelated plans also bear some tangential
similarities to the work in information retrieval on finding similar
or dissimilar documents (c.f. \citep{ir-novelty}).

\section {Conclusion and Future Work}
\label{sec:future}

\noindent
In this paper, we consider the planning problem with partial user's
preference model in two scenarios where the knowledge about preference
is completely unknown or only part of it is given. We propose a general
approach to this problem where a set of plans is presented to the user
from which she can select. For each situation of the incompleteness, we
define different quality measure of plan sets and investigate approaches
to generating plan set with respect to the quality measure. In the first
scenario when the user is known to have preferences over plans, but the
details are completely unknown, we define the quality of plan sets as
their diversity value, specified with syntactic features of plans (its
action set, sequence of states, and set of causal links). We then
consider generating diverse set of plans using two state-of-the-art
planners, {\sc gp-csp} and {\sc lpg}. The approaches we developed for
supporting the generation of diverse plans in {\sc gp-csp} are broadly
applicable to other planners based on bounded horizon compilation
approaches for planning. Similarly, the techniques we developed for {\sc
  lpg}, such as biasing the relaxed plan heuristics in terms of distance
measures, could be applied to other heuristic planners. The experimental
results with {\sc gp-csp} explicate the relative difficulty of enforcing
the various distance measures, as well as the correlation among the
individual distance measures (as assessed in terms of the sets of plans
they find). The experiments with {\sc lpg} demonstrate the potential of
planning using heuristic local search in producing large sets of highly
diverse plans.

When part of the user's preferences is given, in particular the set of
features that the user is interested in and the distribution of weights
representing their relative importance, we propose the usage
of \emph{Integrated Preference Function}, and its special case
\emph{Integrated Convex Preference} function, to measure the quality of plan sets,
and propose various heuristic approaches based on the Metric-LPG
planner~\citep{metric-lpg} to find a good plan set with respect to this
measure. We show empirically that taking partial preferences into
account does improve the quality of plan set returned to the users, and
that our proposed approaches are sensitive to the degree of
preference incompleteness, represented by the distribution.


While a planning agent may well start with a partial preference model,
in the long run, we would like the agent to be able to improve the
preference model through repeated interactions with the user. In our
context, at the beginning when the degree of incompleteness of
preferences is high, the learning will involve improving the estimate of
$h( \alpha )$ based on the feedback about the specific plan that the
user selects from the set returned by the system. This learning phase is
in principle well connected to the Bayesian parameter estimation
approach in the sense that the whole distribution of parameter vector,
$h(\alpha)$, is updated after receiving feedback from the user, taking
into account the current distribution of all models (starting from a
prior, for instance the uniform distribution). Although such interactive
learning framework has been discussed previously, as in
\cite{chajewska2001learning}, the set of user's decisions in this work
is assumed to be given, whereas in planning scenarios the cost of plan
synthesis should be incoporated into the our interactive framework, and
the problem of presenting plan sets to the user needs also to be
considered. Recent work by Li et al. (\citeyear{li2009learning})
considered learning user's preferences in planning, but restricting to 
preference models that can be represented with hierachical task networks.

\medskip
\noindent{\bf Acknowledgements}: We thank Menkes van den Briel for
drawing our attention to ICP measure initially. Tuan Nguyen's research
is supported by a Science Foundation of Arizona fellowship.
Kambhampati's research is supported in part by an IBM Faculty Award, the
NSF grant IIS\u2013308139, ONR grants N00014-09-1-0017,
N00014-07-1-1049, N000140610058, and by a Lockheed Martin subcontract
TT0687680 to ASU as part of the DARPA Integrated Learning program.

\bibliographystyle{elsarticle-harv}
\bibliography{bib.bib}

\end{document}